\documentclass[10pt,journal,compsoc]{IEEEtran}
\ifCLASSOPTIONcompsoc
  \usepackage[nocompress]{cite}
\else
  \usepackage{cite}
\fi
\usepackage{graphicx}
\usepackage{xcolor}
\usepackage{caption}
\usepackage{subcaption}
\usepackage{threeparttable}
\usepackage[numbers]{natbib}
\usepackage{makecell}
\usepackage{booktabs}
\ifCLASSINFOpdf
\else
\fi
\begin{document}
\title{Progressive Cross-stream Cooperation in Spatial and Temporal Domain for Action Localization}

\author{Rui~Su,
        Dong~Xu,~\IEEEmembership{Fellow,~IEEE,}
        Luping~Zhou,~\IEEEmembership{Senior Member,~IEEE,}
        and~Wanli~Ouyang,~\IEEEmembership{Senior Member,~IEEE}% <-this % stops a space
\IEEEcompsocitemizethanks{\IEEEcompsocthanksitem This work was supported by the Australian Research Council (ARC) Future Fellowship under Grant FT180100116 and ARC DP200103223.\hfil\break
\IEEEcompsocthanksitem Rui Su, Dong Xu, Luping Zhou and Wanli Ouyang are with the School of Electrical and Information Engineering, The University of Sydney, NSW, Australia. Dong Xu is the corresponding author. \protect\\
E-mail: rui.su@sydney.edu.au, dong.xu@sydney.edu.au, luping.zhou@sydney.edu.au, wanli.ouyang@sydney.edu.au.
%\IEEEcompsocthanksitem Wen Li is with the Computer Vision Laboratory, ETH Z̈urich, Switzerlan. \hfil\break
%E-mail: liwen@vison.ee.ethz.ch
}% <-this % stops a space% <-this % stops an unwanted space
\thanks{Manuscript received April 19, 2005; revised August 26, 2015.}
}


\markboth{Journal of \LaTeX\ Class Files,~Vol.~14, No.~8, August~2015}%
{Shell \MakeLowercase{\textit{et al.}}: Bare Demo of IEEEtran.cls for Computer Society Journals}



\IEEEtitleabstractindextext{%
\begin{abstract}
Spatio-temporal action localization consists of three levels of tasks: spatial localization, action classification, and temporal localization. In this work, we propose a new Progressive Cross-stream Cooperation (PCSC) framework that improves all three tasks above. The basic idea is to utilize both spatial region (\textit{resp.}, temporal segment proposals) and features from one stream (\textit{i.e.}, the Flow/RGB stream) to help another stream (\textit{i.e.}, the RGB/Flow stream) to iteratively generate better bounding boxes in the spatial domain (\textit{resp.}, temporal segments in the temporal domain). In this way, not only the actions could be more accurately localized both spatially and temporally, but also the action classes could be predicted more precisely.  
Specifically, we first combine the latest region proposals (for spatial detection) or segment proposals (for temporal localization) from both streams to form a larger set of labelled training samples to help learn better action detection or segment detection models. Second, to learn better representations, we also propose a new message passing approach to pass information from one stream to another stream, which also leads to better action detection and segment detection models. By first using our newly proposed PCSC framework for spatial localization at the frame-level and then applying our temporal PCSC framework for temporal localization at the tube-level, the action localization results are progressively improved at both the frame level and the video level. Comprehensive experiments on two benchmark datasets UCF-101-24 and J-HMDB demonstrate the effectiveness of our newly proposed approaches for spatio-temporal action localization in realistic scenarios.

\end{abstract}

\begin{IEEEkeywords}
action localization, spatio-temporal action localization, two-stream cooperation.
\end{IEEEkeywords}}

\maketitle

\IEEEdisplaynontitleabstractindextext

% \fi
\IEEEpeerreviewmaketitle

\IEEEraisesectionheading{\section{Introduction}\label{sec:introduction}}

\IEEEPARstart{R}{ecently}, deep neural networks have significantly improved performance in various computer vision tasks~\cite{he2016deep,ouyang2017deepid,ouyang2017chained,ouyang2013single,sun2018fishnet,liu2018crowd,zhang2018collaborative} including human action detection. Human action detection, also known as the spatio-temporal action localization, aims to recognize the actions of interest presented in videos and localize them in space and time. Due to its wide spectrum of applications, it has attracted increasing research interests. The cluttered background, occlusion and large intra-class variance make spatio-temporal action localization a very challenging task. A brief review of the related works is provided in Section~\ref{Sec:Literature}.

Existing works~\cite{Simonyan_2014_NIPS,Saha_2016_BMVC,peng2016multi} have demonstrated the complementarity between appearance and motion information at the feature level in recognizing and localizing human actions~\cite{Simonyan_2014_NIPS,Saha_2016_BMVC,peng2016multi}. We further observe that these two types of information are also complementary to each other at the region proposal level. That is, either appearance or motion clues alone may succeed or fail to detect region proposals in different scenarios. However, they can help each other by providing region proposals to each other.

For example, the region proposal detectors based on appearance information may fail when certain human actions exhibit extreme poses, but motion information from human movements could help capture human actions. In another example, when the motion clue is noisy because of subtle movement or cluttered background, it is hard to detect positive region proposals based on motion information while the appearance clue may still provide enough information to successfully find candidate action regions and remove a large amount of background or non-action regions. Therefore, we can use the bounding boxes detected from the motion information as the region proposals to improve action detection results based on the appearance information, and vice versa. This underpins our work in this paper to fully exploit the interactions between appearance and motion information at both region proposal and feature levels to improve spatial action localization, which is our first motivation. 

On the other hand, to generate spatio-temporal action localization results, we need to form action tubes by linking the detected bounding boxes for actions in individual frames and temporally segment them out from the entire video clip. Data association methods based on the spatial overlap and action class scores are mainly used in the current works~\cite{Gkioxari_2015_CVPR,peng2016multi,Saha_2016_BMVC, singh2016online}. However, it is difficult for such methods to precisely identify the temporal boundaries of actions. For some cases, due to the subtle difference between the frames near temporal boundaries, it remains an extremely hard challenge to precisely decide the temporal boundary. As a result, it produces a large room for further improvement of the existing temporal refinement methods, which is our second motivation.

Based on the first motivation, in this paper, we propose a progressive framework called Progressive Cross-stream Cooperation (PCSC) to iteratively use both region proposals and features from one stream to progressively help learn better action detection models for another stream. To exploit the information from both streams at the region proposal level, we collect a larger set of training samples by combining the latest region proposals from both streams. At the feature level, we propose a new message passing approach to pass information from one stream to another stream in order to learn better representations. By exploiting the complementality information between appearance and motion clues in this way, we can progressively learn better action detection models and improve action localization results at the frame-level.

Based on the second motivation, we propose a new temporal localization method consisting of an actionness detector and an action segment detector, in which we can extend our PCSC framework to temporal localization. For the actionness detector, we train a set of class-specific binary classifiers (actionness detectors) to detect the happening of a certain type of actions. {These actionness detectors are trained by focusing on ``confusing" samples from the action tube of the same class, and therefore can learn critical features that are good at discriminating the subtle changes across the action boundaries.} Our approach works better when compared with an alternative approach that learns a general actionness detector for all actions. For the action segment detector, we propose a segment proposal based two-stream detector using a multi-branch architecture to address the large temporal scale variation problem. Our PCSC framework can be readily applied to the action segment detector to take advantage of the complementary information between appearance and motion clues to detect accurate temporal boundaries and further improve spatio-temporal localization results at the video-level. 

Our contributions are briefly summarized as follows:
\begin{itemize}
\item For spatial localization, we propose the Progressive Cross-stream Cooperation (PCSC) framework to iteratively use both features and region proposals from one stream to help learn better action detection models for another stream, which includes a new message passing approach and a simple region proposal combination strategy. 
\item We also propose a temporal boundary refinement method to learn class-specific actionness detectors and an action segment detector that applies the newly proposead PCSC framework from the spatial domain to the temporal domain to improve the temporal localization results.
\item Comprehensive experiments on two benchmark datasets UCF-101-24 and J-HMDB demonstrate that our approach outperforms the state-of-the-art methods for localizing human actions both spatially and temporally in realistic scenarios.
\end{itemize}

A preliminary version of our work is published in~\cite{su2019improving}. The difference with~\cite{su2019improving} is summarized as follows. Building upon a Faster-RCNN based action segment detection method for temporal localization~\cite{chao2018rethinking}, we further extend our PCSC framework from the spatial domain to the temporal domain for detecting action segments. Moreover, we integrate the results from the actionness detector proposed in~\cite{su2019improving} and those from our action segment detector discussed in this work to further improve the spatio-temporal localization performance at the video-level. Meanwhile, we provide more analysis for our PCSC framework for action detection in Section ~\ref{analysis} and perform more experiments to evaluate the newly proposed approaches.

\section{Related Work}\label{Sec:Literature}

%--------------------------------------------------------------

\subsection{Spatial temporal localization methods}

Spatio-temporal action localization involves three types of tasks: spatial localization, action classification, and temporal localization. A huge amount of efforts have been dedicated to improve the three tasks from different perspectives. First, for spatial localization, the state-of-the-art human detection methods are utilized to obtain precise object proposals, which includes the use of fast and faster R-CNN in~\cite{Guilhem_2017_IJCVSubmission,Saha_2016_BMVC} as well as Single Shot Multibox Detector (SSD) in~\cite{singh2016online,kalogeiton17iccv}.

Second, discriminant features are also employed for both spatial localization and action classification. For example, some methods~\cite{kalogeiton17iccv,Guilhem_2017_IJCVSubmission} stack neighbouring frames near one key frame to extract more discriminant features in order to remove the ambiguity of actions in each single frame and to better represent this key frame. Other methods~\cite{Singh_2016_CVPR,Pigou_2017_IJCV,Guilhem_2017_IJCVSubmission} utilize recurrent neural networks to link individual frames or use 3D CNNs ~\cite{Carreira_2017_CVPR,Du_ICCV_2015} to exploit temporal information. {Several works~\cite{li2018recurrent,yang2017spatio} detect actions in a single frame and predict action locations in the neighbouring frames to exploit the spatio-temporal consistency.} Meanwhile, complementary information from multi-modalities is also utilized to improve feature extraction results. For example, a number of works~\cite{Saha_2016_BMVC,kalogeiton17iccv,peng2016multi,Guilhem_2017_IJCVSubmission} fuse the appearance and motion clues to extract more robust features for action classification. {Additionally, another work~\cite{zolfaghari2017chained} uses a Markov chain model to integrate the appearance, motion and pose clues for action classification and spatial-temporal action localization.}

Third, temporal localization is to form action tubes from per-frame detection results. The methods for this task are largely based on the association of per-frame detection results, such as the overlap, continuity and smoothness of objects, as well as the action class scores. To improve localization accuracies, a variety of temporal refinement methods have been proposed, e.g., the traditional temporal sliding windows~\cite{peng2016multi}, dynamic programming~\cite{singh2016online, Saha_2016_BMVC}, tubelets linking~\cite{kalogeiton17iccv}, and thresholding-based refinement~\cite{Guilhem_2017_IJCVSubmission}, etc.

Finally, several methods were also proposed to improve action detection efficiency. For example, without requiring time-consuming multi-stage training process, the works in~\cite{Saha_2016_BMVC,peng2016multi} proposed to train a single CNN model by simultaneously performing action classification and bounding box regression. More recently, an online real-time spatio-temporal localization method is also proposed in~\cite{singh2016online}.

%-----------------------------------------------------

\subsection{Two-stream R-CNN}

Based on the observation that appearance and motion clues are often complementary to each other, several state-of-the-art action detection methods~\cite{Saha_2016_BMVC,kalogeiton17iccv,peng2016multi,Guilhem_2017_IJCVSubmission} followed the standard two-stream R-CNN approach. The features extracted from the two streams are fused to improve action detection performance. For example, in~\cite{Saha_2016_BMVC}, the softmax score from each motion bounding box is used to help the appearance bounding boxes with largest overlap. In~\cite{Guilhem_2017_IJCVSubmission}, three types of fusion strategies are discussed: i) simply averaging the softmax outputs of the two streams, ii) learning per-class weights to weigh the original pre-softmax outputs and applying softmax on the weighted sum, and iii) training a fully connected layer on top of the concatenated output from each stream. It is reported in ~\cite{Guilhem_2017_IJCVSubmission} that the third fusion strategy achieves the best performance. {In~\cite{ye2019discovering}, the final detection results are generated by integrating the results from an early-fusion method, which concatenates the appearance and motion clues as the input to generate detection results, and a late-fusion method, which combines the outputs from the two streams.}

Please note that most appearance and motion fusion approaches in the existing works as discussed above are based on the late fusion strategy. They are only trained (if there is any training process) on top of the detection networks of the two streams. In contrast, in this work we iteratively use both types of features and bounding boxes from one stream to progressively help learn better action detection models for another stream, which is intrinsically different with these existing approaches~\cite{singh2016online, kalogeiton17iccv} that fuse two-stream information only at the feature level in a late fusion fashion.

\section{Action Detection Model in Spatial Domain}\label{Sec:ActionDectection}

\begin{centering}
\begin{figure*}[!h]
\begin{center}
\begin{subfigure}[t]{1\textwidth}
\centering
\includegraphics[width=1\textwidth]{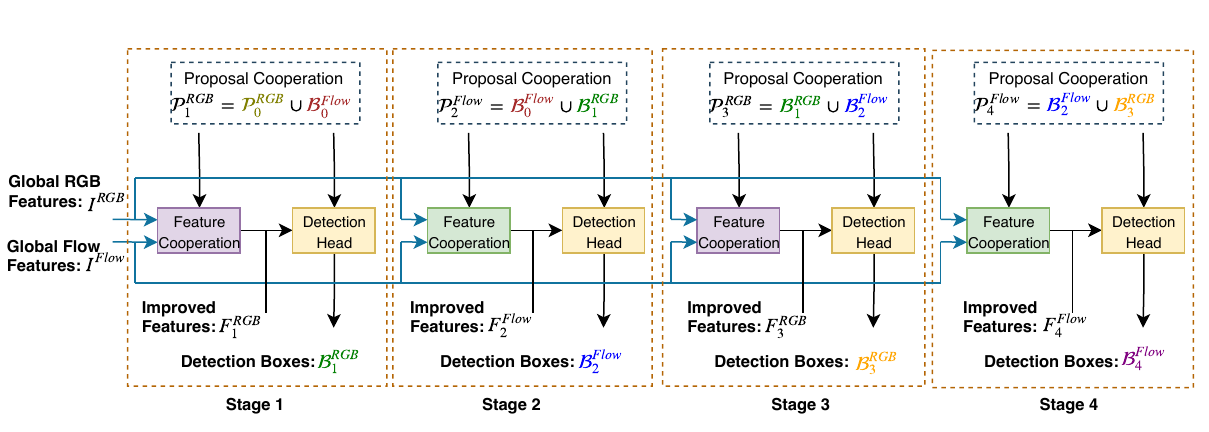}
\caption{}
\end{subfigure}
\begin{subfigure}[t]{0.4\textwidth}
\centering
\includegraphics[width=1\textwidth]{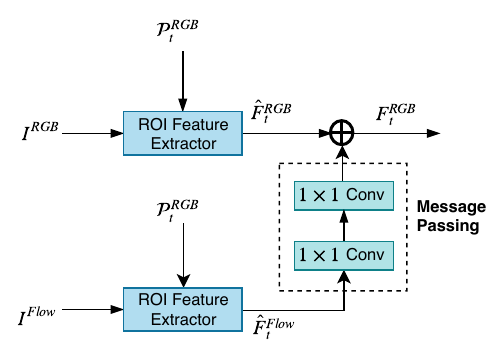}
\caption{}
\end{subfigure}
\begin{subfigure}[t]{0.4\textwidth}
\centering
\includegraphics[width=1\textwidth]{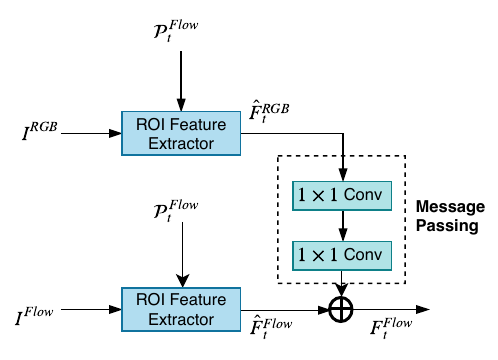}
\caption{}
\end{subfigure}
\caption{(a) Overview of our Cross-stream Cooperation framework (four stages are used as an example for better illustration). The region proposals and features from the flow stream help improve action detection results for the RGB stream at Stage 1 and Stage 3, while the region proposals and features from the RGB stream help improve action detection results for the flow stream at Stage 2 and Stage 4. Each stage comprises of two cooperation modules and the detection head.  The region-proposal level cooperation module refines the region proposals $\mathcal{P}^{i}_{t}$ and the feature-level cooperation module improves the features $\mathbf{F}^{i}_{t}$, where the superscript $i\in\{RGB,Flow\}$ denotes the RGB/Flow stream and the subscript $t$ denotes the stage number.  The detection head is used for estimating the action location and the class label. For region-proposal level cooperation, we combine the most recent region proposals from the two streams. (b) (c) The details of our feature-level cooperation modules through message passing from one stream to another stream. The flow features are used to help the RGB features in (b), while the RGB features are used to help the flow features in (c).
}\label{fig:Overview}
\end{center}
\end{figure*}
\end{centering}

Building upon the two-stream framework~\cite{peng2016multi}, we propose a Progressive Cross-stream Cooperation (PCSC) method for action detection at the frame level. In this method, the RGB (appearance) stream and the flow (motion) stream iteratively help each other at both feature level and region proposal level in  order to achieve better localization results.

\subsection{PCSC Overview}

The overview of our PCSC method (\textit{i.e.}, the frame-level detection method) is provided in Fig.~\ref{fig:Overview}. As shown in Fig.~\ref{fig:Overview} (a), our PCSC is composed of a set of ``stages". Each stage refers to one round of cross-stream cooperation process, in which the features and region proposals from one stream will help improve action localization performance for another stream. Specifically, each stage comprises of two cooperation modules and a detection head module. Our \textbf{detection head} module is a standard one, which consists of several layers for region classification and regression.

The two cooperation modules include region-proposal-level cooperation and feature-level cooperation, which are introduced in details in Section~\ref{subsec:boxes} and Section~\ref{subsec:feats}, respectively. For region-proposal-level cooperation, the detection results from one stream (e.g, the RGB stream) are used as the additional region proposals, which are combined with the region proposals from the other stream (e.g, the flow stream) to refine the region proposals and improve the action localization results. Based on the refined region proposals, we also perform feature-level cooperation by first extracting RGB/flow features from these ROIs and then refining these RGB/flow features via a message-passing module shown in Fig.~\ref{fig:Overview} (b), and Fig.~\ref{fig:Overview} (c), which will be introduced in Section~\ref{subsec:feats}. The refined ROI features in turn lead to better region classification and regression results in the detection head module, which benefits the subsequent action localization process in the next stage. By performing the aforementioned processes for multiple rounds, we can progressively improve the action detection results. The whole network is trained in an end-to-end fashion by minimizing the overall loss, which is the summation of the losses from all stages.

After performing frame-level action detection, our approach links the per-frame detection results to form action tubes, in which the temporal boundary is further refined by using our proposed temporal boundary refinement module. The details are provides in Section ~\ref{sec:tube}.

%---------------------------------------------------------------%
\subsection{Cross-stream Region Proposal Cooperation} \label{subsec:boxes}

We employ the two-stream Faster R-CNN method~\cite{NIPS2015_5638} for frame-level action localization. Each stream has its own Region Proposal Network (RPN)~\cite{NIPS2015_5638} to generate the candidate action regions, and these candidates are then used as the training samples to train a bounding box regression network for action localization. Based on our observation, the region proposals generated by each stream can only partially cover the true action regions, which degrades the detection performance. Therefore, in our method, we use the region proposals from one stream to help another stream.

In this paper, the bounding boxes from RPN are called \textbf{region proposals}, while the bounding boxes from the detection head are called \textbf{detection boxes}.

The region proposals from the RPNs of the two streams are first used to train their own detection head separately in order to obtain their own corresponding detection boxes.
The set of region proposals for each stream is then refined by combining two subsets. The first subset is from the detection boxes of its own stream (e.g, RGB) at the previous stage. The second subset is from the detection boxes of another stream (e.g, flow) at the current stage. To remove more redundant boxes, a lower NMS threshold is used when the detection boxes in another stream (e.g. flow) are used for the current stream (e.g. RGB).

Mathematically, let ${\mathcal P}_t^{(i)}$ and ${\mathcal B}_t^{(i)}$ denote the set of region proposals and the set of the detection boxes, respectively, where $i$ indicates the $i$th stream, $t$ denotes the $t$th stage.
The region proposal ${\mathcal P}_t^{(i)}$ is updated as ${\mathcal P}_t^{(i)} = {\mathcal B}_{t-2}^{(i)} \cup {\mathcal B}_{t-1}^{(j)}$, and the detection box ${\mathcal B}_t^{(j)}$ is updated as ${\mathcal B}_t^{(j)} = { {\mathcal G}}({\mathcal P}_t^{(j)})$, where ${\mathcal G}(\cdot)$ denotes the mapping function from the detection head module. Initially, when $t=0$, $\mathcal{P}_0^{(i)}$ is the region proposal from RPN for the $i$th stream.
This process is repeated between the two streams for several stages, which will progressively improve the detection performance of both streams. 

With our approach, the diversity of region proposals from one stream will be enhanced by using the complementary boxes from another stream. This could help reduce the missing bounding boxes. Moreover, only the bounding boxes with high confidence are utilized in our approach, which increases the chances to add more precisely detected bounding boxes and thus further help improve the detection performance. These aforementioned strategies, together with the cross-modality feature cooperation strategy that will be described in Section~\ref{subsec:feats}, effectively improve the frame-level action detection results. 

Our cross-stream cooperation strategy at the region proposal level shares similar high-level ideas with the two-view learning method co-training ~\cite{Blum:1998:CLU:279943.279962}, as both methods make use of predictions from one stream/view to generate more training samples (\textit{i.e.}, the additional region proposals in our approach) to improve the prediction results for another stream/view. However, our approach is intrinsically different with co-training in the following two aspects. As a semi-supervised learning method, the co-training approach ~\cite{Blum:1998:CLU:279943.279962} selects unlabelled testing samples and assigns pseudo-labels to those selected samples to enlarge the training set. In contrast, the additional region proposals in our work still come from the training videos instead of testing videos, so we know the labels of the new training samples by simply comparing these additional region proposals with the ground-truth bounding boxes. Also, in co-training~\cite{Blum:1998:CLU:279943.279962}, the learnt classifiers will be directly used for predicting the labels of testing data in the testing stage so the complementary information from the two views for the testing data is not exploited. In contrast, in our work, the same pipeline used in the training stage will also be adopted for the testing samples in the testing stage, and thus we can further exploit the complementary information from the two streams for the testing data. 

%-------------------------------------------------%
\subsection{Cross-stream Feature Cooperation} \label{subsec:feats}
To extract spatio-temporal features for action detection, similar to~\cite{Gu_2018_CVPR}, we use the I3D network as the backbone network for each stream. Moreover, we follow Feature Pyramid Network (FPN)~\cite{Lin_2017_CVPR} to build feature pyramid with high-level semantics, which has been demonstrated to be useful to improve bounding box proposals and object detection. This involves a bottom-up pathway and a top-down pathway and lateral connections. The bottom-up pathway uses the feed-forward computation along the backbone I3D network, which produces a feature hierarchy with increasing semantic levels but decreasing spatial resolutions. Then these features are upsampled by using the top-down pathway, which are merged with the corresponding features in the bottom-up pathway through lateral connections.
 
Following~\cite{Lin_2017_CVPR}, we use the feature maps at the layers of Conv2c, Mixed3d, Mixed4f, Mixed5c in I3D to construct the feature hierarchy in the bottom-up pathway, and denote these feature maps as \{$C_2^i$, $C_3^i$, $C_4^i$, $C_5^i$\}, where $i\in\{\mbox{RGB}, \mbox{Flow}\}$ indicates the RGB and the flow streams, respectively. Accordingly, the corresponding feature maps in the top-down pathway are denoted as \{$U_2^i$, $U_3^i$, $U_4^i$, $U_5^i$\}.
 
Most two-stream action detection frameworks~\cite{singh2016online,peng2016multi,kalogeiton17iccv} only exploit the complementary RGB and flow information by fusing softmax scores or concatenating the features before the final classifiers, which are insufficient for the features from the two streams to exchange information from one stream to another and benefit from such information exchange. Based on this observation, we develop a message-passing module to bridge these two streams, so that they help each other for feature refinement. 

We pass the messages between the feature maps in the bottom-up pathway of the two streams. Denote $l$ as the index for the set of feature maps in \{$C_2^i$, $C_3^i$, $C_4^i$, $C_5^i$\}, $l\in\{2, 3, 4, 5\}$. Let us use the improvement of the RGB features as an example (the same method is also applicable to the improvement of the flow features).  Our message-passing module improves the features $C_l^{\mbox{RGB}}$  with the help of the features $C_l^{\mbox{Flow}}$ as follows: 
\begin{equation}\label{Eqn:message-passing}
C_l^{\mbox{RGB}} = f_\theta(C_l^{\mbox{Flow}}) \oplus C_l^{\mbox{RGB}} , 
\end{equation}
where $\oplus$ denotes the element-wise addition of the feature maps, $f_\theta(\cdot)$ is the mapping function (parameterized by $\theta$) of our message-passing module. The function $f_\theta(C_l^{\mbox{Flow}})$ nonlinearly extracts the message from the feature $C_l^{\mbox{Flow}}$, and then uses the extracted message for improving the features $C_l^{\mbox{RGB}}$.

The output of $f_\theta(\cdot)$ has to produce the feature maps with the same number of channels and resolution as $C_l^{\mbox{RGB}}$ and $C_l^{\mbox{Flow}}$. To this end, we design our message-passing module by stacking two $1\times1$ convolutional layers with relu as the activation function. The first $1\times1$ convolutional layer reduces the channel dimension and the second convolutional layer restores the channel dimension back to its original number. This design saves the number of parameters to be learnt in the module and exchanges message by using two-layer non-linear transform. Once the feature maps $C_l^{\mbox{RGB}}$ in the bottom-up pathway are refined, the corresponding features maps $U_l^{\mbox{RGB}}$ in the top-down pathway are generated accordingly.

The above process is for image-level messaging passing only. The image-level message passing is only performed from the Flow stream to the RGB stream once. This message passing provides good features to initialize the message-passing stages.

{Denote the image-level feature map sets for the RGB and flow streams by $I^{\mbox{RGB}}$ and $I^{\mbox{Flow}}$ respectively. They are used to extract features $\hat{F}{}_t^{\mbox{RGB}}$ and $\hat{F}{}_t^{\mbox{Flow}}$ by ROI pooling at each stage $t$, as shown in Fig.~\ref{fig:Overview}. At Stage 1 and Stage 3, the ROI-feature $\hat{F}{}_t^{\mbox{Flow}}$ of the flow stream is used to help improve the ROI-feature $\hat{F}{}_t^{\mbox{RGB}}$ of the RGB stream, as illustrated in Fig.~\ref{fig:Overview} (b). Specifically, the improved RGB feature ${F}{}_t^{\mbox{RGB}}$ is obtained by applying the same method in Eqn.~(\ref{Eqn:message-passing}). Similarly, at Stage 2 and Stage 4, the ROI-feature $\hat{F}{}_t^{\mbox{RGB}}$ of the RGB stream is also used to help improve the ROI-feature $\hat{F}{}_t^{\mbox{Flow}}$ of the flow stream, as illustrated in Fig.~\ref{fig:Overview} (c).
The message passing process between ROI-features aims to provide better features for action box detection and regression, which benefits the next cross-stream cooperation stage.}

\subsection{Training Details} \label{sec:train}

For better spatial localization at the frame-level, we follow~\cite{kalogeiton17iccv} to stack neighbouring frames to exploit temporal context and improve action detection performance for key frames. A key frame is a frame containing the ground-truth actions. Each training sample, which is used to train the RPN in our PCSC method, is composed of $k$ neighbouring frames with the key frame in the middle. The region proposals generated from the RPN are assigned with positive labels when they have an intersection-over-union (IoU) overlap higher than 0.5 with any ground-truth bounding box, or negative labels if their IoU overlap is lower than 0.5 with all ground-truth boxes. This label assignment strategy also applies to the additional bounding boxes from the assistant stream during the region proposal-level cooperation process.

\section{Temporal Boundary Refinement for Action Tubes} \label{sec:tube}

\begin{centering}
\begin{figure*}[!ht]
\begin{center}
\begin{subfigure}[t]{1\textwidth}
\centering
\includegraphics[width=1\textwidth]{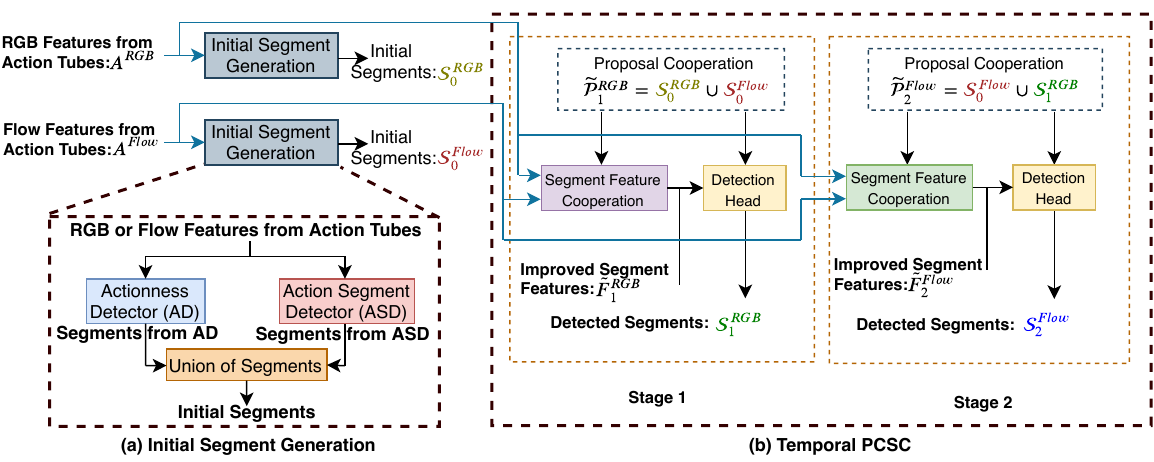}
% \caption{}
\end{subfigure}
% \begin{subfigure}[t]{0.45\textwidth}
% \centering
% \includegraphics[width=1\textwidth]{T-flow2rgb.pdf}
% \caption{}
% \end{subfigure}
% \begin{subfigure}[t]{0.45\textwidth}
% \centering
% \includegraphics[width=1\textwidth]{T-rgb2flow.pdf}
% \caption{}
% \end{subfigure}
\caption{Overview of two modules in our temporal boundary refinement framework: (a) the Initial Segment Generation Module and (b) the Temporal Progressive Cross-stream Cooperation (T-PCSC) Module. The Initial Segment Generation module combines the outputs from Actionness Detector and Action Segment Detector to generate the initial segments for both RGB and flow streams. Our T-PCSC framework takes these initial segments as the segment proposals and iteratively improves the temporal action detection results by leveraging complementary information of two streams at both feature and segment proposal levels (two stages are used as an example for better illustration). The segment proposals and features from the flow stream help improve action segment detection results for the RGB stream at Stage 1, while the segment proposals and features from the RGB stream help improve action detection results for the flow stream at Stage 2. Each stage comprises of the cooperation module and the detection head. The segment proposal level cooperation module refines the segment proposals $\tilde{\mathcal{P}}{}^{i}_{t}$ by combining the most recent segment proposals from the two streams, while the segment feature level cooperation module improves the features $\mathbf{\tilde{F}}{}^{i}_{t}$, where the superscript $i\in\{RGB,Flow\}$ denotes the RGB/Flow stream and the subscript $t$ denotes the stage number. The detection head is used to estimate the action segment location and the actionness probability.
}\label{fig:T-PCSC Overview}
\end{center}
\end{figure*}
\end{centering}

After the frame-level detection results are generated, we then build the action tubes by linking them. Here we use the same linking strategy as in~\cite{singh2016online}, except that we do not apply temporal labeling. Although this linking strategy is robust to missing detection, it is still difficult to accurately determine the start and the end of each action tube, which is a key factor that degrades video-level performance of our action detection framework.

To address this problem, the temporal boundaries of the action tubes need to be refined.{ Our temporal boundary refinement method uses the features extracted from action tubes to refine the temporal boundaries of the action tubes. It is built upon two-stream framework and consists of four modules: Pre-processing Module, Initial Segment Generation module, Temporal Progressive Cross-stream Cooperation (T-PCSC) module, and Post-processing Module. An overview of the two major modules, i.e., the Initial Segment Generation Module and the T-PCSC Module, is provided in Fig.~\ref{fig:T-PCSC Overview}. Specifically,  in the Pre-processing module, we use each bounding box in the action tubes to extract the $7\times7$ feature maps for the detected region on each frame by ROI-Pooling and then temporally stack these feature maps to construct the features for the action tubes. These features are then used as the input to the Initial Segment Generation module. As shown in Fig.~\ref{fig:T-PCSC Overview}, we develop two segment generation methods in the Initial Segment Generation module: (1) a class-specific actionness detector to evaluate actionness (\textit{i.e.} the happening of an action) for each frame, and (2) an action segment detector based on two-stream Faster-RCNN framework to detect action segments. The outputs of the actionness detector and the action segment detector are combined to generate the initial segments for both the RGB and flow streams. Our T-PCSC Module extends our PCSC framework from the spatial domain to the temporal domain for temporal localization, and it uses features from the action tubes and the initial segments generated by the Initial Segment Generation module to encourage the two streams to help each other at both feature level and segment proposal level to improve action segment detection. After multiple stages of cooperation,  the output action segments from each stage are combined in the Post-processing module, in which NMS is also applied to remove redundancy and generate the final action segments.}

\subsection{Actionness Detector (AD)} \label{subsec:act det}

We first develop a class-specific actionness detector to detect the actionness at each given location (spatially and temporally). Taking advantage of the predicted action class labels produced by frame-level detection, we construct our actionness detector by using $N$ binary classifiers, where $N$ is the number of action classes. Each classifier addresses the actionness of a specific class (\textit{i.e.}, whether an action happens or not at a given location in a frame). This strategy is more robust than learning a general actionness classifier for all action classes. {Specifically, after frame-level detection, each bounding box has a class label, based on which, the bounding boxes from the same class are traced and linked to form action tubes~\cite{singh2016online}. To train the binary actionness classifier for each action class $i$, the bounding boxes that are within the action tubes and predicted as class $i$ by the frame-level detector are used as the training samples. Each bounding box is labeled either as $1$ when its overlap with the ground-truth box is greater than 0.5, or as $0$ otherwise. Note that the training samples may contain bounding boxes falsely detected near the temporal boundary of the action tubes. Therefore, these samples are the ``confusing" samples and are useful for the actionness classifier to learn the subtle but critical features, which better determine the beginning and the end of this action. The output of the actionness classifier is a probability of actionness belonging to class $i$.}

At the testing stage, given an action tube formed by using~\cite{singh2016online}, we apply the class-specific actionness detector to every frame-level bounding box in this action tube to predict its actionness probability (called actionness score). Then a median filter over multiple frames is employed to smooth the actionness scores of all bounding boxes in this tube. If a bounding box has a smoothed score lower than a predefined threshold, it will be filtered out from this action tube, and then we can refine the action tubes so that they have more accurate temporal boundaries. Note that when a non-action region near a temporal boundary is falsely detected, it is included in the training set to train our class-specific actionness detector. Therefore, our approach takes advantage of the confusing samples across temporal boundary to obtain better action tubes at the testing stage. 

%

% \begin{centering}
% \begin{figure}[t]
% \begin{center}
% \includegraphics[width=0.5\textwidth]{pami_response_ad.pdf} 
% \caption{\textcolor{red}{Overview of our actionness detector. The action tubes are formed by linking the detected bounding boxes generated by our action detection model at frame level. These action tubes are used to extract features and then the actionness classifiers use these features to predict actionness scores for each frame, forming an anctionness score sequence. A median filter is applied to smooth the actionness score sequence and then the sequence is segmented with a threshold to generate refined segments.}}\label{fig:ad}
% \end{center}
% \end{figure}
% \end{centering}
%
%--------------------------------------------%

\noindent \subsection{Action Segment Detector (ASD)} \label{subsec: TSD}

%
% \begin{centering}
% \begin{figure*}[!h]
% \begin{center}
% \includegraphics[width=0.9\textwidth]{asd} 
% \caption{Overview of our action segment detector, which is based on the PCSC framework in the temporal domain. Our action segment detector without the T-PCSC (ASD w/o T-PCSC) module uses the segment proposal networks (SPN) to generate segment proposals. These segment proposals are then used to extract the segment features from the RGB features from the action tubes by using the segment feature extraction module (same as for flow stream). Then the detection head outputs the detected segments based on the segment features. For our T-PCSC based action segment detector, a two-stream action segment detector is built and the outputs from both streams are used to generate the final segments.}\label{fig:asd}
% \end{center}
% \end{figure*}
% \end{centering}
% %

\begin{centering}
\begin{figure}[t]
\begin{center}
\includegraphics{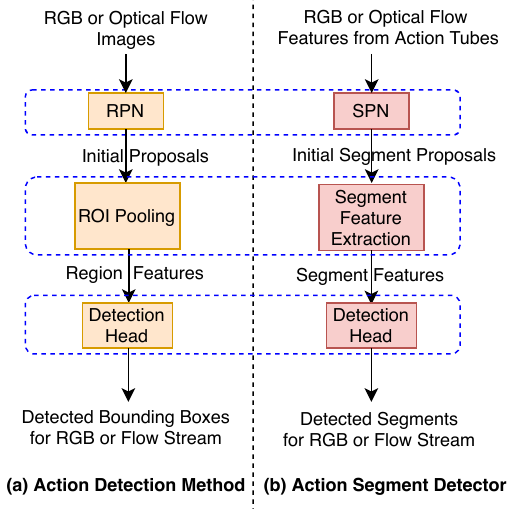} 
\caption{{Comparison of the single-stream frameworks of the action detection method (see Section 3) and our action segment detector (see Section 4.2). The modules within each blue dashed box share similar functions at the spatial domain and the temporal domain, respectively.}}\label{fig:comparison}
\end{center}
\end{figure}
\end{centering}

{{Motivated by the two-stream temporal Faster-RCNN framework~\cite{chao2018rethinking}, we also build an action segment detector, which directly utilizes segment proposals to detect action segments. Our action segment detector takes both RGB and optical flow features from the action tubes as the input, and follows the fast-RCNN framework~\cite{chao2018rethinking} to include a Segment Proposal Network (SPN) for temporal segment proposal generation and  detection heads for segment proposal regression. Specifically, we extract the segment features by using these two-stream features from each action tube and the segment proposals produced by SPN. The segment features are then sent to the corresponding detection head for regression to generate the initial detection of segments. This process is similar to the action detection method described in Section~\ref{Sec:ActionDectection}. A comparison between the single-stream action detection method and the single-stream action segment detector is shown in Figure \ref{fig:comparison}, in which the modules within each blue dashed box share similar functions at the spatial domain and the temporal domain, respectively. Our implementation details for SPN, feature extraction and detection head are introduced below.}}\\

\noindent\textbf{(a) SPN}~~~In Faster-RCNN, the anchors of different object sizes are pre-defined. Similary, in our SPN, $K$ types of action anchors with different duration (time lengths) are pre-defined. The boundaries of the action anchors are regressed by two layers of temporal convolution (called ConvNet in~\cite{chao2018rethinking}) using RGB or flow features from the action tube~\cite{chao2018rethinking}. Note that the object sizes in an image only vary in a relatively small range, but the temporal span of an action can dramatically change in a range from less than one second to more than one minute. Consequently, it is inappropriate for anchors with different scales to share the temporal features from the same receptive field. To address this issue, we use the multi-branch architecture and dilated temporal convolutions in~\cite{chao2018rethinking} for each anchor prediction based on the anchor-specific features. Specifically, for each scale of anchors, a sub-network with the same architecture but different dilation rate is employed to generate features with different receptive fields and decide the temporal boundary. In this way, our SPN consists of $K$ sub-networks targeting at different time lengths. In order to consider the context information, for each anchor with the scale $s$, we require the receptive field to cover the context information with temporal span $s/2$ before and after the start and the end of the anchor. We then regress temporal boundaries and predict actionness probabilities for the anchors by applying two parallel temporal convolutional layers with the kernel size of 1. The output segment proposals are ranked according to their proposal actionness probabilities and NMS is applied to remove the redundant proposals. Please refer to~\cite{chao2018rethinking} for more details.\\

\noindent\textbf{(b) Segment feature extraction}~~~
After generating segment proposals, we construct segment features to refine the temporal boundaries and predict the segment actionness probabilities. For each segment proposal with the length $l_s$, the start point $t_{start}$ and the end point $t_{end}$, we extract three types of features, $F_{center}$, $F_{start}$, and $F_{end}$. We use temporal ROI-Pooling on top of the segment proposal regions to pool action tube features and obtain the features $F_{center} \in R^{D\times l_t}$, where $l_t$ is the temporal length and $D$ is the dimension of the features at each temporal location. 
As the context information immediately before and after the segment proposal is helpful to decide the temporal boundaries of an action, this information should be represented by our segment features. Therefore, we additionally use temporal ROI-Pooling to pool features $F_{start}$ (\textit{resp.}, $F_{end}$), from the segments centered at $t_{start}$ (\textit{resp.}, $t_{end}$) with the length of $2l_s/5$ to represent the starting (reps., ending) information. We then concatenate $F_{start}$, $F_{center}$ and $F_{end}$ to construct the segment features $F_{seg}$. \\

%Moreover, since the length $l_s$ could be very different across segment proposals, $F_{seg}$ with fixed $l_t$ cannot well capture the boundary information when $l_s$ is large. 
%To address this issue, we quantize all segment proposals into $M$ groups with the length range $R_m=[r_m,r_{m+1}]$, $m=1,2,...,M$, where $r_m$ and $r_{m+1}$ are the lower and the upper bounds of the segment length in the $m$-th group. Accordingly, the pooled feature length $l_t$ for each group is denoted as $l_m$ where $m=1,2,...,M$. That is, when the length $l_s$ is in the range $[r_m, r_m+1]$, then the features for the segment proposal regions will be pooled to features with $l_t=l_m$.

% \begin{centering}
% \begin{figure}[!h]
% \begin{center}

% \includegraphics[width=0.5\textwidth]{TSB} 

% \caption{Illustration of the temporal supervision branch. The input features of this branch are also used as the input features of the classification and the regression branches.}\label{fig:tsb}

% \end{center}
% \end{figure}
% \end{centering}

\noindent\textbf{(c) Detection head}~~~The extracted segment features are passed to the detection head to obtain the refined segment boundaries and the segment actionness probabilities. As the temporal length of the extracted temporal features is not fixed, similar to SPN, the detection head module consists of $M$ sub-networks corresponding to the $M$ groups of the temporal length of the segment proposals resulting from SPN. For each sub-network, a spatial fully-connected layer is first used to reduce the spatial resolution, and then two temporal convolution layers with the kernel size of 3 are applied to non-linearly combine temporal information. The output features are used to predict the actionness probabilities and regress the segment boundaries. For actionness classification, in addition to minimize the segment-level actionness loss, we also minimize the frame-level actionness loss. We observe that it is beneficial to extract discriminative features by predicting actionness for each frame, which will eventually help refine the boundaries of segments.

\subsection{Temporal PCSC (T-PCSC)}
{Our PCSC framework can leverage the complementary information at the feature level and the region proposal level to iteratively improve action detection performance at the spatial domain. Similarly, we propose a Temporal PCSC (T-PCSC) module that extends our PCSC framework from the spatial domain to the temporal domain for temporal localization. Specifically, we combine the outputs generated by the actionness detector and the action segment detector to obtain the initial segments for each stream. These initial segments are then used by our T-PCSC module as the initial segment proposals to generate action segments.}

 ~~~Similar to our PCSC method described in Section~\ref{Sec:ActionDectection}, our T-PCSC method consists of the feature cooperation module and proposal cooperation module, as shown in Fig.~\ref{fig:T-PCSC Overview}. Note that for the message-passing block in the feature cooperation module, the two $1 \times 1$ convolution layers are replaced by two temporal convolution layers with the kernel size of 1. We only apply T-PCSC for two stages in our action segment detector method due to memory constraints and the observation that we already achieve good results after using two stages. Similar to our PCSC framework shown in Figure~\ref{fig:Overview}, in the first stage of our T-PCSC framework, the initial segments from the RGB stream are combined with those from the flow stream to form a new segment proposal set by using the proposal cooperation module. This new segment proposal set is used to construct the segment features as described in Section~\ref{subsec: TSD} (b) for both the RGB and the flow streams, and then the feature cooperation module passes the messages from the flow features to the RGB features to refine the RGB features. The refined RGB features are then used to generate the output segments by the detection head described in Section~\ref{subsec: TSD} (c). A similar process takes place in the second stage. But this time, the output segments from the flow stream and the first stage are used to form the segment proposal set and the message passing process is conducted from the RGB features to the flow features to help refine the flow features. We then generate the final action segments by combining the output action segments of each stage and applying NMS to remove redundancy and obtain the final results.\\

\section{Experimental results}

We introduce our experimental setup and datasets in Section~\ref{exp-setup}, and then compare our method with the state-of-the-art methods in Section~\ref{comparison}, and conduct ablation study in Section~\ref{ablation}.

%-------------------------------
\subsection{Experimental Setup}\label{exp-setup}

\textbf{Datasets.}~~We evaluate our PCSC method on two benchmarks: UCF-101-24~\cite{soomro2012ucf101} and J-HMDB-21~\cite{jhuang2013towards}. \textbf{UCF-101-24} contains 3207 untrimmed videos from 24 sports classes, which is a subset of the UCF-101 dataset, with spatio-temporal annotations provided by~\cite{singh2016online}. Following the common practice, we use the predefined ``split 1" protocol to split the training and test sets, and report the results based on this split.  \textbf{J-HMDB-21} contains 928 videos from 21 action classes. All the videos are trimmed to contain the actions only. We perform the experiments on three predefined training-test splits, and report the average results on this dataset.

\textbf{Metrics.}~~We evaluate the action detection performance at both frame-level and video-level by mean Average precision (mAP). To calculate mAP, a detection box is considered as a correct one when its overlap with a ground-truth box or tube is greater than a threshold $\delta$. The overlap between our detection results and the ground truth is measured by the intersection-over-union (IoU) score at the frame-level and the spatio-temporal tube overlap at the video-level, respectively. In addition, we also report the results based on the COCO evaluation metric~\cite{lin2014microsoft}, which averages the mAPs over 10 different IoU thresholds from 0.5 to 0.95 with the interval of 0.05. 

To evaluate the localization accuracy of our method, we also calculate Mean Average Best Overlap (MABO). We compute the IoU scores between the bounding boxes from our method and the ground-truth boxes, and then average the IoU scores from the bounding boxes with largest overlap to the ground-truth boxes for each class. The MABO is calculated by averaging the averaged IoU scores over all classes.

To evaluate the classification performance of our method, we report the average confident score over different IoU threshold $\theta$. The average confident score is computed by averaging the confident scores of the bounding boxes from our method that have an IoU with the ground-truth boxes greater than $\theta$.

\textbf{Implementation Details.} We use the I3D features~\cite{Carreira_2017_CVPR} for both streams, and the I3D model is pretrained with Kinetics. The optical flow images are extracted by using FlowNet v2~\cite{ilg2017flownet}. The mini-batch size used to train the RPN and the detection heads in our action detection method is 256 and 512, respectively. Our PCSC based action detection method is trained for 6 epochs by using three 1080Ti GPUs. The initial learning rate is set as 0.01, which drops $10\%$ at the 5th epoch and another $10\%$ at the 6th epoch. {For our actionness detector, we set the window size as 17 when using median filter to smooth the actionness scores. we also use a predefined threshold to decide the starting and ending locations for action tubes, which is empirically set as 0.01.} For segment generation, we apply the anchors with the following $K$ scales ($K=11$ in this work): \{6, 12, 24, 42, 48, 60, 84, 92, 144, 180, 228\}. The length ranges $R$, which are used to group the segment proposals, are set as follows: $\{[20, 80], [80, 160], [160, 320], [320, \infty]\}$, \textit{i.e.}, we have $M=4$, and the corresponding feature lengths $l_t$ for the length ranges are set to $\{15, 30, 60, 120\}$. {{The numbers of the scales and the lengths are empirically decided based on the frame numbers.}} {In this work, we apply NMS to remove redundancy and combine the detection results from different stages with an overlap threshold of 0.5.} The mini-batch sizes used to train the SPN and the detection heads in our temporal refinement method are set as 128 and 32, respectively. Our PCSC based temporal refinement method is trained for 5 epochs and the initial learning rate is set as 0.001, which drops $10\%$ at the 4th epoch and another $10\%$ at the 5th epoch.

\subsection{Comparison with the State-of-the-art methods}\label{comparison}

We compare our PCSC method with the state-of-the-art methods. The results of the existing methods are quoted directly from their original papers. In addition, we also evaluate the object detection method (denoted as ``Faster R-CNN + FPN") in~\cite{Lin_2017_CVPR} with the I3D network as its backbone network, in which we report the results by only using the single stream (\textit{i.e.}, RGB or Flow) features as well as the result by applying NMS to the union set of the two stream bounding boxes. Specifically, the only difference between the method in~\cite{Lin_2017_CVPR} and our PCSC is that our PCSC method has the cross-stream cooperation modules while the work in~\cite{Lin_2017_CVPR} does not have them. Therefore, the comparison between the two approaches can better demonstrate the benefit of our cross-stream cooperation strategy.

\subsubsection{Results on the UCF-101-24 Dataset}
The results on the UCF-101-24 dataset are reported in Table~\ref{tab:UCF-frame} and Table~\ref{tab:UCF-video}. 

\begin{table}[h!]
    \centering
    \caption{Comparison (mAPs \% at the frame level) of different methods on the UCF-101-24 dataset when using the IoU threshold $\delta$ at 0.5.}
    \begin{threeparttable}
    \begin{tabular}{l c c}
    \hline
         & Streams & mAPs \\
        \hline
        \citet{Weinzaepfel_2015_ICCV} & RGB+Flow & 35.8 \\
        \citet{peng2016multi} & RGB+Flow & 65.7 \\
        \citet{ye2019discovering}& RGB+Flow & 67.0 \\
        \citet{kalogeiton17iccv} & RGB+Flow & 69.5 \\
        \citet{Gu_2018_CVPR} & RGB+Flow& 76.3 \\
        \hline
        Faster R-CNN + FPN (RGB)~\cite{Lin_2017_CVPR}&RGB & 73.2 \\
        Faster R-CNN + FPN (Flow)~\cite{Lin_2017_CVPR}&Flow & 64.7 \\
        Faster R-CNN + FPN~\cite{Lin_2017_CVPR}&RGB+Flow & 75.5 \\
        PCSC (Ours)& RGB+Flow & \textbf{79.2} \\
    \hline
    \end{tabular}
    \end{threeparttable}
    \label{tab:UCF-frame}
\end{table}

\begin{table}[h!]
\centering
\caption{Comparison (mAPs \% at the video level) of different methods on the UCF-101-24 dataset when using different IoU thresholds. Here ``AD" is for actionness detector, and ``ASD" is for action segment detector.}
\begin{threeparttable}
\begin{tabular}{l c c c c c} 
 \hline
 IoU threshold $\delta$ & Streams & 0.2 & 0.5 & 0.75 & 0.5:0.95\\ 
 \hline
 \citet{Weinzaepfel_2015_ICCV} & RGB+Flow & 46.8 & - & - & - \\
 \citet{zolfaghari2017chained} & \makecell{RGB+Flow\\ +Pose} & 47.6 & -& -& -\\
 \citet{peng2016multi} & RGB+Flow& 73.5 & 32.1 & 02.7 & 07.3 \\
 \citet{Saha_2016_BMVC} &RGB+Flow& 66.6 & 36.4 & 0.8 & 14.4 \\
 \citet{yang2017spatio}&RGB+Flow&73.5 & 37.8& -& -\\
 \citet{singh2016online}&RGB+Flow & 73.5 & 46.3 & 15.0 & 20.4 \\
 \citet{ye2019discovering}&RGB+Flow&76.2& -& -& -\\
 \citet{kalogeiton17iccv}&RGB+Flow & 76.5 & 49.2 & 19.7 & 23.4 \\
 \citet{li2018recurrent}&RGB+Flow&77.9&-&-&- \\
 \citet{Gu_2018_CVPR} &RGB+Flow & - & 59.9 & - & -  \\
 \hline
 \makecell[l]{Faster R-CNN \\+ FPN (RGB)~\cite{Lin_2017_CVPR}} & RGB & 77.5 & 51.3 & 14.2 & 21.9 \\ 
 \makecell[l]{Faster R-CNN \\+ FPN (Flow)~\cite{Lin_2017_CVPR}} & Flow & 72.7 & 44.2 & 7.4 & 17.2 \\ 
 \makecell[l]{Faster R-CNN \\+ FPN~\cite{Lin_2017_CVPR}} & RGB+Flow & 80.1 & 53.2 & 15.9 & 23.7 \\ 
 PCSC + AD~\cite{su2019improving} & RGB+Flow& 84.3 & 61.0 & 23.0 & 27.8 \\
 \makecell[l]{PCSC + ASD+ AD \\+ T-PCSC (Ours)}& RGB+Flow & \textbf{84.7} & \textbf{63.1} & \textbf{23.6} & \textbf{28.9} \\
 \hline
\end{tabular}
\end{threeparttable}
\label{tab:UCF-video}
\end{table}

Table~\ref{tab:UCF-frame} shows the mAPs from different methods at the frame-level on the UCF-101-24 dataset. All mAPs are calculated based on the IoU threshold $\delta = 0.5$. From Table~\ref{tab:UCF-frame}, we observe that our PCSC method achieves an mAP of $79.2\%$, outperforming all the existing methods by a large margin. Especially, our PCSC performs better than Faster R-CNN + FPN~\cite{Lin_2017_CVPR} by an improvement of $3.7$\%. This improvement is fully due to the proposed cross-stream cooperation framework, which is the only difference between ~\cite{Lin_2017_CVPR} and our PCSC. It is interesting to observe that both methods~\cite{kalogeiton17iccv} and~~\cite{Gu_2018_CVPR} additionally utilize the temporal context for per frame-level action detection. However, both methods do not fully exploit the complementary information between the appearance and motion clues, and therefore they are worse than our method. Moreover, our method also outperforms~\cite{kalogeiton17iccv} and~\cite{Gu_2018_CVPR} by $9.7$\% and $2.9$\%, respectively, in terms of frame-level mAPs.

Table~\ref{tab:UCF-video} reports the video-level mAPs at various IoU thresholds (0.2, 0.5, and 0.75) on the UCF-101-24 dataset. The results based on the COCO evaluation metrics~\cite{lin2014microsoft} are reported in the last column of Table~\ref{tab:UCF-video}. Our method is denoted as ``PCSC + ASD + AD + T-PCSC" in Table~\ref{tab:UCF-video}, where the proposed temporal boundary refinement method, consisting of three modules (\textit{i.e.}, the actionness detector module, the action segment detector module, and the Temporal PCSC), is applied to refine the action tubes generated from our PCSC method. Our preliminary work~\cite{su2019improving}, which only uses the actionness detector module as the temporal boundary refinement method, is denoted as ''PCSC + AD". Consistent with the observations at the frame-level, our method  outperforms all the state-of-the-art methods under all evaluation metrics. When using the IoU threshold $\delta = 0.5$, we achieve an mAP of $63.1\%$ on the UCF-101-24 dataset. This result beats~\cite{kalogeiton17iccv} and~\cite{Gu_2018_CVPR}, which only achieve the mAPs of 49.2\% and 59.9\%, respectively. Moreover, as the IoU threshold increases, the performance of our method drops less when compared with other state-of-the-art methods. The results demonstrate that our detection method achieves higher localization accuracy than other competitive methods.

\subsubsection{Results on the J-HMDB Dataset}

For the J-HMDB dataset, the frame-level mAPs and the video-level mAPs from different mothods are reported in Table~\ref{tab:HMDB-frame} and Table~\ref{tab:HMDB-video}, respectively. Since the videos in the J-HMDB dataset are trimmed to only contain actions, the temporal refinement process is not required, so we do not apply our temporal boundary refinement module when generating action tubes on the J-HMDB dataset. 

\begin{table}[h!]
    \centering
    \caption{Comparison (mAPs \% at the frame level) of different methods on the J-HMDB dataset when using the IoU threshold $\delta$ at 0.5.}
    \begin{threeparttable}
    \begin{tabular}{l c c}
    \hline
         & Streams & mAPs \\
        \hline
        \citet{peng2016multi} &RGB+Flow & 58.5 \\
        \citet{ye2019discovering} & RGB+Flow& 63.2 \\
        \citet{kalogeiton17iccv} & RGB+Flow& 65.7 \\
        \citet{Hou_2017_ICCV} &RGB& 61.3 \\
        \citet{Gu_2018_CVPR} &RGB+Flow& 73.3 \\
        \citet{Sun_2018_ECCV} &RGB+Flow& \textbf{77.9} \\
        \hline
        Faster R-CNN + FPN (RGB)~\cite{Lin_2017_CVPR} &RGB & 64.7 \\
        Faster R-CNN + FPN (Flow)~\cite{Lin_2017_CVPR} &Flow & 68.4 \\
        Faster R-CNN + FPN~\cite{Lin_2017_CVPR} &RGB+Flow & 70.2 \\
        PCSC (Ours)&RGB+Flow & 74.8 \\
    \hline
    \end{tabular}
    \end{threeparttable}
    \label{tab:HMDB-frame}
\end{table}
\begin{table}[h!]
    \centering
    \caption{Comparison (mAPs \% at the video level) of different methods on the J-HMDB dataset when using different IoU thresholds.}
    \begin{tabular}{l c c c c c}
    \hline
        IoU threshold $\delta$& Streams & 0.2 & 0.5 & 0.75 & 0.5:0.95 \\
        \hline
        \citet{Gkioxari_2015_CVPR} & RGB+Flow & - & 53.3 & - & -  \\
        \citet{Wang_2016_CVPR} & RGB+Flow & - & 56.4 & - & - \\
        \citet{Weinzaepfel_2015_ICCV}& RGB+Flow & 63.1 & 60.7 & - & - \\
        \citet{Saha_2016_BMVC}& RGB+Flow & 72.6 & 71.5 & 43.3 & 40.0 \\
        \citet{peng2016multi} & RGB+Flow & 74.1 & 73.1 & - & - \\
        \citet{singh2016online}& RGB+Flow & 73.8 & 72.0 & 44.5 & 41.6 \\
        \citet{kalogeiton17iccv}& RGB+Flow & 74.2 & 73.7 & 52.1 & 44.8 \\
        \citet{ye2019discovering}& RGB+Flow & 75.8 & 73.8& - & - \\
        \citet{zolfaghari2017chained}&\makecell{RGB+Flow\\ +Pose}& 78.2 & 73.4 & -& -\\
        \citet{Hou_2017_ICCV} & RGB& 78.4 & 76.9 & - & - \\
        \citet{Gu_2018_CVPR} & RGB+Flow& - & 78.6 & - & - \\
        \citet{Sun_2018_ECCV} & RGB+Flow& - & 80.1 & - & -\\
        \hline
        \makecell[l]{Faster R-CNN \\+ FPN (RGB)~\cite{Lin_2017_CVPR}}& RGB & 74.5 & 73.9 & 54.1 & 44.2 \\
        \makecell[l]{Faster R-CNN \\+ FPN (Flow)~\cite{Lin_2017_CVPR}}& Flow & 77.9 & 76.1 & 56.1 & 46.3 \\
        \makecell[l]{Faster R-CNN \\+ FPN~\cite{Lin_2017_CVPR}}& RGB+Flow & 79.1 & 78.5 & 57.2 & 47.6 \\
        PCSC (Ours) & RGB+Flow & \textbf{82.6} & \textbf{82.2} & \textbf{63.1} & \textbf{52.8} \\
    \hline
    \end{tabular}
    \label{tab:HMDB-video}
\end{table}

We have similar observations as in the UCF-101-24 dataset. At the video-level, our method is again the best performer under all evaluation metrics on the J-HMDB dataset (see Table~\ref{tab:HMDB-video}). When using the IoU threshold $\delta = 0.5$, our PCSC method outperforms~\cite{Gu_2018_CVPR} and \cite{Sun_2018_ECCV} by 3.6\% and 2.1\%, respectively. 

At the frame-level (see Table~\ref{tab:HMDB-frame}), our PCSC method performs the second best, which is only worse than a very recent work~\cite{Sun_2018_ECCV}. However, the work in \cite{Sun_2018_ECCV} uses S3D-G as the backbone network, which provides much stronger features when compared with the I3D features used in our method. In addition, please note that, our method outperforms~\cite{Sun_2018_ECCV} in terms of mAPs at the video level (see Table~\ref{tab:HMDB-video}), which demonstrates promising performance of our PCSC method. Moreover, as a general framework, our PCSC method could also take advantage of strong features provided by the S3D-G method to further improve the results, which will be explored in our future work.

\subsection{Ablation Study}\label{ablation}

In this section, we take the UCF-101-24 dataset as an example to investigate the contributions of different components in our proposed method.

\begin{table}[h!]
    \centering
    \caption{Ablation study for our PCSC method at different training stages on the UCF-101-24 dataset.}
    \begin{tabular}{ccc }
    \hline
        Stage & \makecell{PCSC w/o feature \\ cooperation} & \makecell{PCSC }  \\
         \hline
        0 & 75.5 & \textbf{78.1} \\
        1 & 76.1 & \textbf{78.6} \\
        2 & 76.4 & \textbf{78.8} \\
        3 & 76.7 &\textbf{79.1} \\
        4 & 76.7 & \textbf{79.2} \\
    \hline
    \end{tabular}
    \label{tab:4}
\end{table}

\begin{table}[h!]
    \centering
    \caption{Ablation study for our temporal boundary refinement method on the UCF-101-24 dataset. Here ``AD" is for the actionness detector while ``ASD" is for the action segment detector. }
    \begin{tabular}{l c}
    \hline
         & Video mAP (\%) \\
         \hline
        Faster R-CNN + FPN~\cite{Lin_2017_CVPR} & 53.2 \\
        PCSC & 56.4 \\
        PCSC + AD & 61.0 \\
        PCSC + ASD (single branch) & 58.4\\
        PCSC + ASD  & 60.7 \\
        PCSC + ASD + AD & 61.9 \\
        PCSC + ASD + AD + T-PCSC & \textbf{63.1} \\
    \hline
    \end{tabular}
    \label{tab:5}
\end{table}

% \begin{table}[h!]
%     \centering
%     \caption{Ablation study for the multi-branch architecture in our segment proposal network (SPN) and detection head on the UCF-101-24 dataset. Here ``SB" refers to single-branch, ``MB" refers to multi-branch.}
%     \begin{tabular}{c c c}
%     \hline
%          SPN & Detection head& Video mAP (\%) \\
%          \hline
%         SB & SB & 58.9 \\
%         MB & SB & 59.9 \\
%         MB & MB & \textbf{60.7} \\
%     \hline
%     \end{tabular}
%     \label{tab:multi-branch-analysis}
% \end{table}

\begin{table}[h!]
    \centering
    \caption{Ablation study for our Temporal PCSC (T-PCSC) method at different training stages on the UCF-101-24 dataset.}
    \begin{tabular}{cc }
    \hline
        Stage & \makecell{Video mAP (\%)}  \\
         \hline
        0 & {61.9} \\
        1 & {62.8} \\
        2 & \textbf{63.1} \\
    \hline
    \end{tabular}
    \label{tab:t-pcsc-stages}
\end{table}

\textbf{Progressive cross-stream cooperation.}~~In Table~\ref{tab:4}, we report the results of an alternative approach of our PCSC (called PCSC w/o feature cooperation). In the second column of Table~\ref{tab:4}, we remove the feature-level cooperation module from Fig.~\ref{fig:Overview}, and only use the region-proposal-level cooperation module. As our PCSC is conducted in a progressive manner, we also report the performance at different stages to verify the benefit of this progressive strategy. It is worth mentioning that the output at each stage is obtained by combining the detected bounding boxes from both the current stage and all previous stages, and then we apply non-maximum suppression (NMS) to obtain the final bounding boxes. For example, the output at Stage 4 is obtained by applying NMS to the union set of the detected bounding boxes from Stages 0, 1, 2, 3 and 4. At Stage 0, the detection results from both RGB and the Flow streams are simply combined. From Table~\ref{tab:4}, we observe that the detection performance of our PCSC method with or without the feature-level cooperation module is improved as the number of stages increases. However, such improvement seems to become saturated when reaching Stage 4, as indicated by the marginal performance gain from Stage 3 to Stage 4. Meanwhile, when comparing our PCSC with the alternative approach PCSC w/o feature cooperation at every stage, we observe that both region-proposal-level and feature-level cooperation contributes to performance improvement.

% \begin{centering}

% \begin{figure}[t]

% \begin{center}

% \includegraphics[width=0.4\textwidth]{LoR} 

% \caption{The LoR of the region proposals from the latest RGB stream with respect to the region proposals from the latest flow stream at every stage in our PCSC action detector model on the UCF-101-24 dataset.}\label{fig:LoR}

% \end{center}

% \end{figure}

% \end{centering}

\textbf{Action tubes refinement.} {In Table~\ref{tab:5}, we investigate the effectiveness of our temporal boundary refinement method by switching on and off the key components in our temporal boundary refinement module, in which video-level mAPs at the IoU threshold $\delta=0.5$ are reported. In Table~\ref{tab:5}, we investigate {five} configurations for temporal boundary refinement. Specifically, in `PCSC + AD", we use the PCSC method for spatial localization together with the actionness detector (AD) for temporal localization.  In ``PCSC + ASD", we use the PCSC method for spatial localization, while we use the action segment detection method for temporal localization. {{In ``PCSC + ASD (single branch)", we use the ``PCSC + ASD" configuration, but replace the multi-branch architecture in the action segment detection method with the single-branch architecture.}} In ``PCSC + ASD + AD", the PCSC method is used for spatial localization together with both AD and ASD for temporal localization. In ``PCSC + ASD + AD + T-PCSC", the PCSC method is used for spatial localization and both actionness detector and action segment detector are used to generate the initial segments, and our T-PCSC is finally used for temporal localization. 
%our PCSC model with the actionness detector (AD) is denoted as ``PCSC + AD". We also evaluate our action segment detection model without temporal PCSC module (ASD w/o PCSC), which is denoted as ``PCSC + ASD w/o T-PCSC". We also report the results by applying both AD and ASD w/o T-PCSC, which is referred to as ``PCSC + AD + ASD w/o T-PCSC". 
By generating higher quality detection results at each frame, our PCSC method in the spatial domain outperforms the work in~\cite{Lin_2017_CVPR} that does not use the two-stream cooperation strategy in the spatial domain by 3.2\%. After applying our actionness detector module to further boost the video-level performance, we arrive at the video-level mAP of 61.0\%. Meanwhile, our action segment detector without using T-PCSC has already been able to achieve the video-level mAP of 60.7\%, which is comparable to the performance from the actionness detector module in terms of video-level mAPs. {{On the other hand, the video-level mAP drops 2.3\% after replacing the multi-branch architecture in the action segment detector with the single-branch architecture, which demonstrates the effectiveness of the multi-branch architecture.}} Note that the video-level mAP is further improved when combining the results from actionness detector and the action segment detector without using the temporal PCSC module, which indicates that the frame-level and the segment-level actionness results could be complementary. When we further apply our temporal PCSC framework, the video-level mAP of our proposed method increases 1.2\%, and arrives at the best performance of 63.1\%, which demonstrates the effectiveness of our PCSC framework for temporal refinement. In total, we achieve about 7.0\% improvement after applying our temporal boundary refinement module on top of our PCSC method, which indicates the necessity and benefits of the temporal boundary refinement process in determining the temporal boundaries of action tubes.}

%\textbf{Multi-branch Architecture.} We investigate the effectiveness of the multi-branch architecture in our SPN and detection head. In table~\ref{tab:multi-branch-analysis}, the video-level mAPs at the IoU threshold $\delta=0.5$ are reported for our SPN and detection head with/without multi-branch architecture in the action segment detector without temporal PCSC model. We observe that the model can achieve video-level mAP at 58.9\% when applying single-branch architecture for both the SPN and the detection head. After applying multi-branch architecture for the SPN, the video-levle mAP of the model increases to 59.9\%. Moreover, further 0.8\% improvement is observed when applying mulit-branch architecture for both the SPN and the detection head. These results demonstrate the effectiveness of the multi-branch architecture for both our SPN and detection head.

\begin{table}[t]
    \centering
    \caption{The large overlap ratio (LoR) of the region proposals from the latest RGB stream  with respective to the region proposals from the latest flow stream at each stage in our PCSC action detector method on the UCF-101-24 dataset.}
    \begin{tabular}{c c}
    \hline
         Stage &  LoR(\%) \\
         \hline
        1 & 61.1\\
        2 & 71.9\\
        3 & 95.2\\
        4 & \textbf{95.4}\\
    \hline
    \end{tabular}
    \label{tab:LoR}
\end{table}

\begin{table}[h!]
    \centering
    \caption{The Mean Average Best Overlap (MABO) of the bounding boxes with the ground-truth boxes at each stage in our PCSC action detector method on the UCF-101-24 dataset.}
    \begin{tabular}{c c}
    \hline
         Stage & MABO(\%) \\
         \hline
        1 & 84.1\\
        2 & 84.3\\
        3 & 84.5\\
        4 & \textbf{84.6}\\
    \hline
    \end{tabular}
    \label{tab:6}
\end{table}

\begin{table}[h!]
    \centering
    \caption{The average confident scores (\%) of the bounding boxes with respect to the IoU threshold $\theta$ at different stages on the UCF-101-24 dataset.}
    \begin{tabular}{c c c c c}
    \hline
         $\theta$ & Stage 1 &  Stage 2 & Stage 3 & Stage 4 \\
         \hline
        0.5 & 55.3 & 55.7 & 56.3 & \textbf{57.0}\\
        0.6 & 62.1 & 63.2 & 64.3 & \textbf{65.5}\\
        0.7 & 68 & 68.5 & 69.4 & \textbf{69.5}\\
        0.8 & 73.9 & 74.5 & 74.8 & \textbf{75.1}\\
        0.9 & 78.4 & 78.4 & 79.5 & \textbf{80.1}\\
    \hline
    \end{tabular}
    \label{tab:7}
\end{table}

\begin{table}[!h]
\setlength\tabcolsep{3pt}\small
\caption{Detection speed in frame per second (FPS) for our action detection module in the spatial domain on UCF-101-24 when using different number of stages. }
\begin{center}
% \resizebox{1.0\linewidth}{!}{%
\begin{tabular}{c c c c c c c }
\toprule
Stage & 0 & 1 & 2 & 3 & 4\\
\midrule
Detection speed (FPS) &3.1&2.9&2.8&2.6&2.4 \\
\bottomrule
\end{tabular}
% }
\end{center}
\label{table:runtime-fps}
\end{table}

\begin{table}[!h]
\setlength\tabcolsep{3pt}\small
\caption{Detection speed in video per second (VPS) for our temporal boundary refinement method for action tubes on UCF-101-24 when using different number of stages. }
\begin{center}
% \resizebox{1.0\linewidth}{!}{%
\begin{tabular}{c c c c c c c }
\toprule
Stage & 0 & 1 & 2\\
\midrule
Detection speed (VPS) &2.1&1.7&1.5 \\
\bottomrule
\end{tabular}
% }
\end{center}
\label{table:runtime-asd}
\end{table}

\textbf{Temporal progressive cross-stream cooperation.} In Table~\ref{tab:t-pcsc-stages}, we report the video-level mAPs at the IoU threshold $\delta=0.5$ by using our temporal PCSC method at different stages to verify the benefit of this progressive strategy in the temporal domain. Stage 0 is our ``PCSC + ASD + AD" method. Stage 1 and Stage 2 refer to our ``PCSC + ASD + AD + T-PCSC" method using one-stage and two-stage temproal PCSC, respectively. The video-level performance after using our temporal PCSC method is improved as the number of stages increases, which demonstrates the effectiveness of the progressive strategy in the temporal domain.

%===================================================

% \begin{centering}

% \begin{figure}[t]

% \begin{center}

% \includegraphics[width=0.4\textwidth]{MABO} 

% \caption{The MABO of the bounding boxes with the ground-truth boxes at every stage on the UCF-101-24 dataset.}\label{fig:MABO}

% \end{center}

% \end{figure}

% \end{centering}

\subsection{Analysis of Our PCSC at Different Stages}\label{analysis}

{In this section, we take the UCF-101-24 dataset as an example to further investigate our PCSC framework for spatial localization at different stages.}

{In order to investigate the similarity change of proposals generated from two streams over multiple stages, for each proposal from the RGB stream, we compute its maximum IoU (mIoU) with respect to all the proposals from the flow stream. We define the large overlap ratio (LoR) as the percentage of the RGB-stream proposals whose mIoU with all the flow-stream proposals is larger than $0.7$ over the total number of RGB proposals. Table~\ref{tab:LoR} shows the LoR of the latest RGB stream with respect to the latest flow stream at every stage on the UCF-101-24 dataset. Specifically, at stage 1 or stage 3, LoR indicates the overlap degree of the RGB-stream proposals generated at the current stage (i.e., stage 1 or stage 3, respectively) with respect to the flow-stream proposals generated at the previous stage (i.e., stage 0 or stage 2, respectively);  at stage 2 or stage 4, LoR indicates the overlap degree of the RGB-stream proposals generated at the previous stage (i.e., stage 1 or stage 3, respectively) with respect to the flow-stream proposals generated at the current stage (i.e., stage 2 or stage 4, respectively). We observe that the LoR increases from 61.1\% at stage 1 to 95.4\% at stage 4, which demonstrates that the similarity of the region proposals from these two streams increases over stages. At stage 4, the region proposals from different streams are very similar to each other, and therefore, these two types of proposals are lack of complementary information. As a result, there is no further improvement after stage 4.}
%refers to the LoR of the region proposals generated at stage 1 with respect to the region proposals generated from the flow stream at stage 0 (before proceeding to our PCSC), while stage 2 refers to the LoR of the region proposals generated at stage 1 with respect to the region proposals generated at stage 2. We observe that the LoR increases from 61.1\% at stage 1 to 95.4\% at stage 4, which demonstrates that the similarity of the region proposals from these two streams increases over stages. At stage 4, the region proposals from different streams are very similar to each other, and therefore, these two types of proposals are lack of complementary information. As a result, there is no further improvement after stage 4.}

% \begin{centering}

% \begin{figure}[t]

% \begin{center}

% \includegraphics[width=0.4\textwidth]{confident_scores.pdf} 

% \caption{The average confident scores of the bounding boxes with respect to the IoU threshold $\theta$ at different stage on the UCF-101-24 dataset.}\label{fig:CS}

% \end{center}

% \end{figure}

% \end{centering}

{To further evaluate the improvement of our PCSC method over stages, we report MABO and the average confident score to measure the localization and classification performance of our PCSC method over stages. Table~\ref{tab:6} and Table~\ref{tab:7} show the MABO and the average confident score of our PCSC method at every stages on the UCF-101-24 dataset. In terms of both metrics, improvement after each stage can be observed clearly, which demonstrates the effectiveness of our PCSC method for both localization and classification.}

\begin{centering}

\begin{figure}[t]

\begin{center}

\begin{tabular}{cc} \small
&\\
\multicolumn{2}{c}{\includegraphics{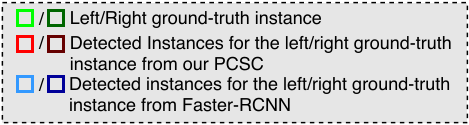}}\\
\includegraphics[width=0.2\textwidth]{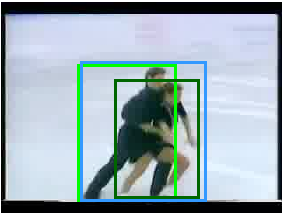} &

\includegraphics[width=0.2\textwidth]{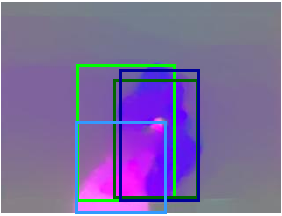} \\

\makecell{(a)} & \makecell{(b)}\\

\includegraphics[width=0.2\textwidth]{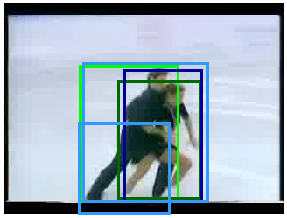} &

\includegraphics[width=0.2\textwidth]{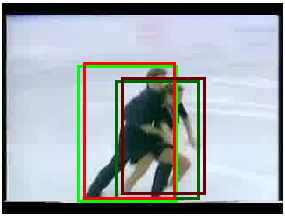}\\

\makecell{(c)}. & \makecell{(d)}\\

\end{tabular}
\caption{An example of visualization results from different methods for one frame. (a) single-stream Faster-RCNN (RGB stream), (b) single-stream Faster-RCNN (flow stream), (c) a simple combination of two-stream results from Faster-RCNN~\cite{Lin_2017_CVPR}, and (d) our PCSC. (Best viewed on screen.)}\label{fig:visual_models}

\end{center}

\end{figure}

\end{centering}

\begin{centering}

\begin{figure}[t]

\begin{center}

\begin{tabular}{cc} \small
&\\

\multicolumn{2}{c}{\includegraphics{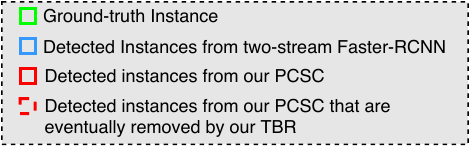}}\\

\includegraphics[width=0.2\textwidth]{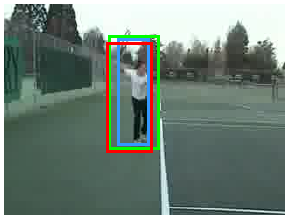} &

\includegraphics[width=0.2\textwidth]{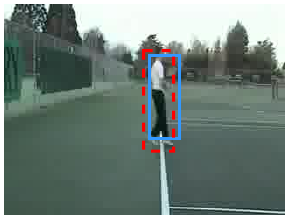} \\

\makecell{(a)} & \makecell{(b)}\\

\includegraphics[width=0.2\textwidth]{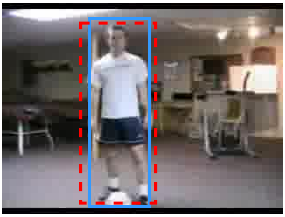} &

\includegraphics[width=0.2\textwidth]{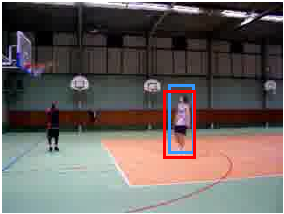} \\

\makecell{(c)} & \makecell{(d)}\\
\end{tabular}

\caption{Example results from our PCSC with temporal boundary refinement (TBR) and the two-stream Faster-RCNN method. There is one ground-truth instance in (a) and no ground-truth instance in (b-d). In (a), both of our PCSC with TBR method and the two-stream Faster-RCNN method~\cite{Lin_2017_CVPR} successfully capture the ground-truth action instance ``Tennis Swing". In (b) and (c), both of our PCSC method and the two-stream Faster-RCNN method falsely detect the bounding boxes with the class label ``Tennis Swing" for (b) and ``Soccer Juggling" for (c). However, our TBR method can remove the falsely detected bounding boxes from our PCSC in both (b) and (c). In (d), both of our PCSC method and the two-stream Faster-RCNN method falsely detect the action bounding boxes with the class label ``Basketball" and our TBR method cannot remove the falsely detected bounding box from our PCSC in this failure case. (Best viewed on screen.)}\label{fig:visual_AD}

% \caption{Example results from our action detection method with temporal refinement by using our actionness detector and a two-stream Faster-RCNN method. In (a), both of our action detection method and the two-stream Faster-RCNN successfully capture the ground-truth action instance ``Tennis Swing". In (b) and (c), although there is no ground-truth action instance in these frames, both of our action detection method and the two-stream Faster-RCNN falsely detect bounding boxes with class label ``Tennis Swing" for (b) and ``Soccer Juggling" for (c) due to the similar scenes and motion patterns to the corresponding foreground frames. However, by temporal refinement, our actionness detector can assign low actionness scores to the falsely detected bounding boxes by our action detection method so that they can be removed. In (d), both of our action detection method and the two-stream Faster-RCNN falsely detect action bounding boxes while there are no ground-truth action instances in these frames. Although our actionness detector can reduce the actionness scores for these falsely detected bounding boxes by our action detection method, these scores are yet not sufficiently low to make the falsely detected bounding boxes removed. (Best viewed on screen.)}\label{fig:visual_AD}

\end{center}

\end{figure}

\end{centering}

\subsection{Runtime Analysis}

{We report the detection speed of our action detection method in spatial domain and our temporal boundary refinement method for detecting action tubes when using different number of stages and the results are provided in Table \ref{table:runtime-fps} and Table \ref{table:runtime-asd}, respectively. We observe that the detection speed of both our action detection method in the spatial domain and the temporal boundary refinement method for detecting action tubes slightly decreases as the number of stages increases (note that Stage 0 in Table~\ref{table:runtime-fps} corresponds to the baseline method Faster-RCNN~\cite{Lin_2017_CVPR} and Stage 0 in Table~\ref{table:runtime-asd} corresponds to the results from our Initial Segment Generation module). Therefore, there is a trade-off between the localization performance and the speed by using the optimal number of stages.}

\subsection{Qualitative Results}

{In addition to the quantitative performance comparison, we also provide some visualization results to further demonstrate the performance of our method. In Fig. \ref{fig:visual_models}, we show the visualization results generated by using either single RGB or flow stream, simple combination of two streams and our two-stream cooperation method. The detection method using only the RGB or flow stream leads to either missing detection (see Fig. \ref{fig:visual_models} (a)) or false detection (see Fig. \ref{fig:visual_models} (b)). As shown in Fig.~\ref{fig:visual_models} (c), a simple combination of these two results can solve the missing detection problem but cannot avoid the false detection. In contrast, our action detection method can refine the detected bounding boxes produced by either stream and remove some falsely detected bounding boxes (see Fig. \ref{fig:visual_models} (d)). 

Moreover, the effectiveness of our temporal refinement method is also demonstrated in Figure \ref{fig:visual_AD}. We compare the results from our method and the two-stream Faster-RCNN method~\cite{Lin_2017_CVPR}. Although both our PCSC method and the two-stream Faster-RCNN method can correctly detect the ground-truth action instance (see Fig. \ref{fig:visual_AD} (a)), after the refinement process by using our temporal boundary refinement module (TBR), the actionness score for the correctly detected bounding box from our PCSC is further increased. On the other hand, because of the similar scene and motion patterns in the foreground frame (see Fig. \ref{fig:visual_AD} (a)) and the background frame (see Fig. \ref{fig:visual_AD} (b)), both our PCSC method and the two-stream Faster-RCNN method generate bounding boxes in the background frame, although there is no ground-truth action instance. However, for the falsely detected bounding box from our PCSC method, our temporal boundary refinement (TBR) module can significantly decrease the actionness score from $0.74$ to $0.04$, which eventually removes the falsely detected bounding box. The similar observation can be found in Fig.~\ref{fig:visual_AD} (c).

% \begin{centering}

% \begin{figure}[!h]

% \begin{center}

% \begin{tabular}{cc} \small

% \includegraphics[width=0.2\textwidth]{pami_response_3a.pdf} &

% \includegraphics[width=0.2\textwidth]{pami_response_3b.pdf} \\

% \makecell{(a)} & \makecell{(b)}
% \end{tabular}

% \caption{Failure cases of our method and a two-stream Faster-RCNN method. The red boxes indicate falsely detected bounding boxes by our action detection method, which can not be removed even after applying our actionness detector. The pink boxes indicate falsely detected bounding boxes by the two-stream Faster-RCNN which are not removed. Both of our action detection method and the two-stream Faster-RCNN falsely detect action bounding boxes while there are no ground-truth action instances in these frames. Although our actionness detector can reduce the actionness scores for these falsely detected bounding boxes by our action detection method, these scores are yet not sufficiently low to make the falsely detected bounding boxes removed. (Best viewed on screen.)}\label{fig:visual_error}

% \end{center}

% \end{figure}

% \end{centering}

% 	We also provide some visual examples of failure cases to show the limitations of our model. As shown in Figure \ref{fig:visual_error}, in most failure cases, our model falsely generated bounding boxes for the background frames, which do not have any ground-truth action instances. 
	While our method can remove some falsely detected bounding boxes as shown in Fig. \ref{fig:visual_AD} (b) and Fig.~\ref{fig:visual_AD} (c), there are still some challenging cases that our method cannot work (see Fig.~\ref{fig:visual_AD} (d)). Although the actionness score of the falsely detected action instance could be reduced after the temporal refinement process by using our TBR module, it could not be completely removed due to the relatively high score. We also observe that the state-of-the-art methods (e.g., two-stream Faster-RCNN~\cite{Lin_2017_CVPR}) cannot work well for this challenging case, either. Therefore, how to reduce false detection from the background frames is still an open issue, which will be studied in our future work.}

\section{Conclusion}

In this work, we have proposed the Progressive Cross-stream Cooperation (PCSC) framework to improve the spatio-temporal action localization results, in which we first use our PCSC framework for spatial localization at the frame level and then apply our temporal PCSC framework for temporal localization at the action tube level. Our PCSC framework consists of several iterative stages. At each stage, we progressively improve action localization results for one stream (\textit{i.e.}, RGB/flow) by leveraging the information from another stream (\textit{i.e.}, RGB/flow) at both region proposal level and feature level. The effectiveness of our newly proposed approaches is demonstrated by extensive experiments on both UCF-101-24 and J-HMDB datasets.

% or


% \appendices

% % you can choose not to have a title for an appendix
% % if you want by leaving the argument blank
% \section{}

% % use section* for acknowledgment
% \ifCLASSOPTIONcompsoc
%   % The Computer Society usually uses the plural form
%   \section*{Acknowledgments}
% \else
%   % regular IEEE prefers the singular form
% \fi


\ifCLASSOPTIONcaptionsoff
  \newpage
\fi



%\bibliographystyle{IEEEtran}


% \end{thebibliography}

{\small
\bibliography{egbib}

\begin{thebibliography}{36}
\providecommand{\natexlab}[1]{#1}
\providecommand{\url}[1]{\texttt{#1}}
\expandafter\ifx\csname urlstyle\endcsname\relax
  \providecommand{\doi}[1]{doi: #1}\else
  \providecommand{\doi}{doi: \begingroup \urlstyle{rm}\Url}\fi

\bibitem[Blum and Mitchell(1998)]{Blum:1998:CLU:279943.279962}
A.~Blum and T.~Mitchell.
\newblock Combining labeled and unlabeled data with co-training.
\newblock In \emph{Proceedings of the Eleventh Annual Conference on
  Computational Learning Theory}, COLT' 98, pages 92--100, New York, NY, USA,
  1998. ACM.
\newblock ISBN 1-58113-057-0.
\newblock \doi{10.1145/279943.279962}.
\newblock URL \url{http://doi.acm.org/10.1145/279943.279962}.

\bibitem[Carreira and Zisserman(2017)]{Carreira_2017_CVPR}
J.~Carreira and A.~Zisserman.
\newblock Quo vadis, action recognition? a new model and the kinetics dataset.
\newblock In \emph{The IEEE Conference on Computer Vision and Pattern
  Recognition (CVPR)}, July 2017.

\bibitem[Chao et~al.(2018)Chao, Vijayanarasimhan, Seybold, Ross, Deng, and
  Sukthankar]{chao2018rethinking}
Y.-W. Chao, S.~Vijayanarasimhan, B.~Seybold, D.~A. Ross, J.~Deng, and
  R.~Sukthankar.
\newblock Rethinking the faster r-cnn architecture for temporal action
  localization.
\newblock In \emph{Proceedings of the IEEE Conference on Computer Vision and
  Pattern Recognition}, pages 1130--1139, 2018.

\bibitem[Ch{\'{e}}ron et~al.(2018)Ch{\'{e}}ron, Osokin, Laptev, and
  Schmid]{Guilhem_2017_IJCVSubmission}
G.~Ch{\'{e}}ron, A.~Osokin, I.~Laptev, and C.~Schmid.
\newblock Modeling spatio-temporal human track structure for action
  localization.
\newblock \emph{CoRR}, abs/1806.11008, 2018.
\newblock URL \url{http://arxiv.org/abs/1806.11008}.

\bibitem[Gkioxari and Malik(2015)]{Gkioxari_2015_CVPR}
G.~Gkioxari and J.~Malik.
\newblock Finding action tubes.
\newblock In \emph{The IEEE Conference on Computer Vision and Pattern
  Recognition (CVPR)}, 2015.

\bibitem[Gu et~al.(2018)Gu, Sun, Ross, Vondrick, Pantofaru, Li,
  Vijayanarasimhan, Toderici, Ricco, Sukthankar, Schmid, and
  Malik]{Gu_2018_CVPR}
C.~Gu, C.~Sun, D.~A. Ross, C.~Vondrick, C.~Pantofaru, Y.~Li,
  S.~Vijayanarasimhan, G.~Toderici, S.~Ricco, R.~Sukthankar, C.~Schmid, and
  J.~Malik.
\newblock Ava: A video dataset of spatio-temporally localized atomic visual
  actions.
\newblock In \emph{The IEEE Conference on Computer Vision and Pattern
  Recognition (CVPR)}, June 2018.

\bibitem[He et~al.(2016)He, Zhang, Ren, and Sun]{he2016deep}
K.~He, X.~Zhang, S.~Ren, and J.~Sun.
\newblock Deep residual learning for image recognition.
\newblock In \emph{Proceedings of the IEEE conference on computer vision and
  pattern recognition}, pages 770--778, 2016.

\bibitem[Hou et~al.(2017)Hou, Chen, and Shah]{Hou_2017_ICCV}
R.~Hou, C.~Chen, and M.~Shah.
\newblock Tube convolutional neural network (t-cnn) for action detection in
  videos.
\newblock In \emph{The IEEE International Conference on Computer Vision
  (ICCV)}, Oct 2017.

\bibitem[Ilg et~al.(2017)Ilg, Mayer, Saikia, Keuper, Dosovitskiy, and
  Brox]{ilg2017flownet}
E.~Ilg, N.~Mayer, T.~Saikia, M.~Keuper, A.~Dosovitskiy, and T.~Brox.
\newblock Flownet 2.0: Evolution of optical flow estimation with deep networks.
\newblock In \emph{IEEE conference on computer vision and pattern recognition
  (CVPR)}, volume~2, page~6, 2017.

\bibitem[Jhuang et~al.(2013)Jhuang, Gall, Zuffi, Schmid, and
  Black]{jhuang2013towards}
H.~Jhuang, J.~Gall, S.~Zuffi, C.~Schmid, and M.~J. Black.
\newblock Towards understanding action recognition.
\newblock In \emph{Proceedings of the IEEE international conference on computer
  vision}, pages 3192--3199, 2013.

\bibitem[Kalogeiton et~al.(2017)Kalogeiton, Weinzaepfel, Ferrari, and
  Schmid]{kalogeiton17iccv}
V.~Kalogeiton, P.~Weinzaepfel, V.~Ferrari, and C.~Schmid.
\newblock Action tubelet detector for spatio-temporal action localization.
\newblock In \emph{ICCV}, 2017.

\bibitem[Li et~al.(2018)Li, Qiu, Dai, Yao, and Mei]{li2018recurrent}
D.~Li, Z.~Qiu, Q.~Dai, T.~Yao, and T.~Mei.
\newblock Recurrent tubelet proposal and recognition networks for action
  detection.
\newblock In \emph{Proceedings of the European conference on computer vision
  (ECCV)}, pages 303--318, 2018.

\bibitem[Lin et~al.(2014)Lin, Maire, Belongie, Hays, Perona, Ramanan,
  Doll{\'a}r, and Zitnick]{lin2014microsoft}
T.-Y. Lin, M.~Maire, S.~Belongie, J.~Hays, P.~Perona, D.~Ramanan,
  P.~Doll{\'a}r, and C.~L. Zitnick.
\newblock Microsoft coco: Common objects in context.
\newblock In \emph{European conference on computer vision}, pages 740--755.
  Springer, 2014.

\bibitem[Lin et~al.(2017)Lin, Dollar, Girshick, He, Hariharan, and
  Belongie]{Lin_2017_CVPR}
T.-Y. Lin, P.~Dollar, R.~Girshick, K.~He, B.~Hariharan, and S.~Belongie.
\newblock Feature pyramid networks for object detection.
\newblock In \emph{The IEEE Conference on Computer Vision and Pattern
  Recognition (CVPR)}, July 2017.

\bibitem[Liu et~al.(2018)Liu, Wang, Li, Ouyang, and Lin]{liu2018crowd}
L.~Liu, H.~Wang, G.~Li, W.~Ouyang, and L.~Lin.
\newblock Crowd counting using deep recurrent spatial-aware network.
\newblock In \emph{Proceedings of the 27th International Joint Conference on
  Artificial Intelligence}, pages 849--855. AAAI Press, 2018.

\bibitem[Ouyang and Wang(2013)]{ouyang2013single}
W.~Ouyang and X.~Wang.
\newblock Single-pedestrian detection aided by multi-pedestrian detection.
\newblock In \emph{Proceedings of the IEEE Conference on Computer Vision and
  Pattern Recognition}, pages 3198--3205, 2013.

\bibitem[Ouyang et~al.(2017{\natexlab{a}})Ouyang, Wang, Zhu, and
  Wang]{ouyang2017chained}
W.~Ouyang, K.~Wang, X.~Zhu, and X.~Wang.
\newblock Chained cascade network for object detection.
\newblock In \emph{Proceedings of the IEEE International Conference on Computer
  Vision}, pages 1938--1946, 2017{\natexlab{a}}.

\bibitem[Ouyang et~al.(2017{\natexlab{b}})Ouyang, Zeng, Wang, Qiu, Luo, Tian,
  Li, Yang, Wang, Li, et~al.]{ouyang2017deepid}
W.~Ouyang, X.~Zeng, X.~Wang, S.~Qiu, P.~Luo, Y.~Tian, H.~Li, S.~Yang, Z.~Wang,
  H.~Li, et~al.
\newblock Deepid-net: Object detection with deformable part based convolutional
  neural networks.
\newblock \emph{IEEE Transactions on Pattern Analysis and Machine
  Intelligence}, 39\penalty0 (7):\penalty0 1320--1334, 2017{\natexlab{b}}.

\bibitem[Peng and Schmid(2016)]{peng2016multi}
X.~Peng and C.~Schmid.
\newblock Multi-region two-stream r-cnn for action detection.
\newblock In \emph{European Conference on Computer Vision}, pages 744--759.
  Springer, 2016.

\bibitem[Pigou et~al.(2018)Pigou, van~den Oord, Dieleman, Herreweghe, and
  Dambre]{Pigou_2017_IJCV}
L.~Pigou, A.~van~den Oord, S.~Dieleman, M.~V. Herreweghe, and J.~Dambre.
\newblock Beyond temporal pooling: Recurrence and temporal convolutions for
  gesture recognition in video.
\newblock \emph{International Journal of Computer Vision}, 126\penalty0
  (2-4):\penalty0 430--439, 2018.

\bibitem[Ren et~al.(2015)Ren, He, Girshick, and Sun]{NIPS2015_5638}
S.~Ren, K.~He, R.~Girshick, and J.~Sun.
\newblock Faster r-cnn: Towards real-time object detection with region proposal
  networks.
\newblock In \emph{Advances in Neural Information Processing Systems 28}, pages
  91--99. Curran Associates, Inc., 2015.

\bibitem[Saha et~al.(2016)Saha, Singh, Sapienza, Torr, and
  Cuzzolin]{Saha_2016_BMVC}
S.~Saha, G.~Singh, M.~Sapienza, P.~H.~S. Torr, and F.~Cuzzolin.
\newblock Deep learning for detecting multiple space-time action tubes in
  videos.
\newblock In \emph{Proceedings of the British Machine Vision Conference 2016,
  {BMVC} 2016, York, UK, September 19-22, 2016}, 2016.

\bibitem[Simonyan and Zisserman(2014)]{Simonyan_2014_NIPS}
K.~Simonyan and A.~Zisserman.
\newblock Two-stream convolutional networks for action recognition in videos.
\newblock In \emph{Advances in Neural Information Processing Systems 27}, pages
  568--576, 2014.

\bibitem[Singh et~al.(2016)Singh, Marks, Jones, Tuzel, and
  Shao]{Singh_2016_CVPR}
B.~Singh, T.~K. Marks, M.~Jones, O.~Tuzel, and M.~Shao.
\newblock A multi-stream bi-directional recurrent neural network for
  fine-grained action detection.
\newblock In \emph{The IEEE Conference on Computer Vision and Pattern
  Recognition (CVPR)}, June 2016.

\bibitem[Singh et~al.(2017)Singh, Saha, Sapienza, Torr, and
  Cuzzolin]{singh2016online}
G.~Singh, S.~Saha, M.~Sapienza, P.~Torr, and F.~Cuzzolin.
\newblock Online real time multiple spatiotemporal action localisation and
  prediction.
\newblock 2017.

\bibitem[Soomro et~al.(2012)Soomro, Zamir, and Shah]{soomro2012ucf101}
K.~Soomro, A.~R. Zamir, and M.~Shah.
\newblock Ucf101: A dataset of 101 human actions classes from videos in the
  wild.
\newblock 2012.

\bibitem[Su et~al.(2019)Su, Ouyang, Zhou, and Xu]{su2019improving}
R.~Su, W.~Ouyang, L.~Zhou, and D.~Xu.
\newblock Improving action localization by progressive cross-stream
  cooperation.
\newblock In \emph{Proceedings of the IEEE Conference on Computer Vision and
  Pattern Recognition}, pages 12016--12025, 2019.

\bibitem[Sun et~al.(2018{\natexlab{a}})Sun, Shrivastava, Vondrick, Murphy,
  Sukthankar, and Schmid]{Sun_2018_ECCV}
C.~Sun, A.~Shrivastava, C.~Vondrick, K.~Murphy, R.~Sukthankar, and C.~Schmid.
\newblock Actor-centric relation network.
\newblock In \emph{The European Conference on Computer Vision (ECCV)},
  September 2018{\natexlab{a}}.

\bibitem[Sun et~al.(2018{\natexlab{b}})Sun, Pang, Shi, Yi, and
  Ouyang]{sun2018fishnet}
S.~Sun, J.~Pang, J.~Shi, S.~Yi, and W.~Ouyang.
\newblock Fishnet: A versatile backbone for image, region, and pixel level
  prediction.
\newblock In \emph{Advances in Neural Information Processing Systems}, pages
  762--772, 2018{\natexlab{b}}.

\bibitem[Tran et~al.(2015)Tran, Bourdev, Fergus, Torresani, and
  Paluri]{Du_ICCV_2015}
D.~Tran, L.~Bourdev, R.~Fergus, L.~Torresani, and M.~Paluri.
\newblock Learning spatiotemporal features with 3d convolutional networks.
\newblock In \emph{Proceedings of the 2015 IEEE International Conference on
  Computer Vision (ICCV)}, ICCV '15, 2015.

\bibitem[Wang et~al.(2016)Wang, Qiao, Tang, and Van~Gool]{Wang_2016_CVPR}
L.~Wang, Y.~Qiao, X.~Tang, and L.~Van~Gool.
\newblock Actionness estimation using hybrid fully convolutional networks.
\newblock In \emph{The IEEE Conference on Computer Vision and Pattern
  Recognition (CVPR)}, June 2016.

\bibitem[Weinzaepfel et~al.(2015)Weinzaepfel, Harchaoui, and
  Schmid]{Weinzaepfel_2015_ICCV}
P.~Weinzaepfel, Z.~Harchaoui, and C.~Schmid.
\newblock Learning to track for spatio-temporal action localization.
\newblock In \emph{The IEEE International Conference on Computer Vision
  (ICCV)}, December 2015.

\bibitem[Yang et~al.(2017)Yang, Gao, and Nevatia]{yang2017spatio}
Z.~Yang, J.~Gao, and R.~Nevatia.
\newblock Spatio-temporal action detection with cascade proposal and location
  anticipation.
\newblock \emph{arXiv preprint arXiv:1708.00042}, 2017.

\bibitem[Ye et~al.(2019)Ye, Yang, and Tian]{ye2019discovering}
Y.~Ye, X.~Yang, and Y.~Tian.
\newblock Discovering spatio-temporal action tubes.
\newblock \emph{Journal of Visual Communication and Image Representation},
  58:\penalty0 515--524, 2019.

\bibitem[Zhang et~al.(2018)Zhang, Ouyang, Li, and Xu]{zhang2018collaborative}
W.~Zhang, W.~Ouyang, W.~Li, and D.~Xu.
\newblock Collaborative and adversarial network for unsupervised domain
  adaptation.
\newblock In \emph{Proceedings of the IEEE Conference on Computer Vision and
  Pattern Recognition}, pages 3801--3809, 2018.

\bibitem[Zolfaghari et~al.(2017)Zolfaghari, Oliveira, Sedaghat, and
  Brox]{zolfaghari2017chained}
M.~Zolfaghari, G.~L. Oliveira, N.~Sedaghat, and T.~Brox.
\newblock Chained multi-stream networks exploiting pose, motion, and appearance
  for action classification and detection.
\newblock In \emph{Proceedings of the IEEE International Conference on Computer
  Vision}, pages 2904--2913, 2017.

\end{thebibliography}
}

\begin{IEEEbiography}
[{\includegraphics[width=1in,height=1.25in,clip,keepaspectratio]{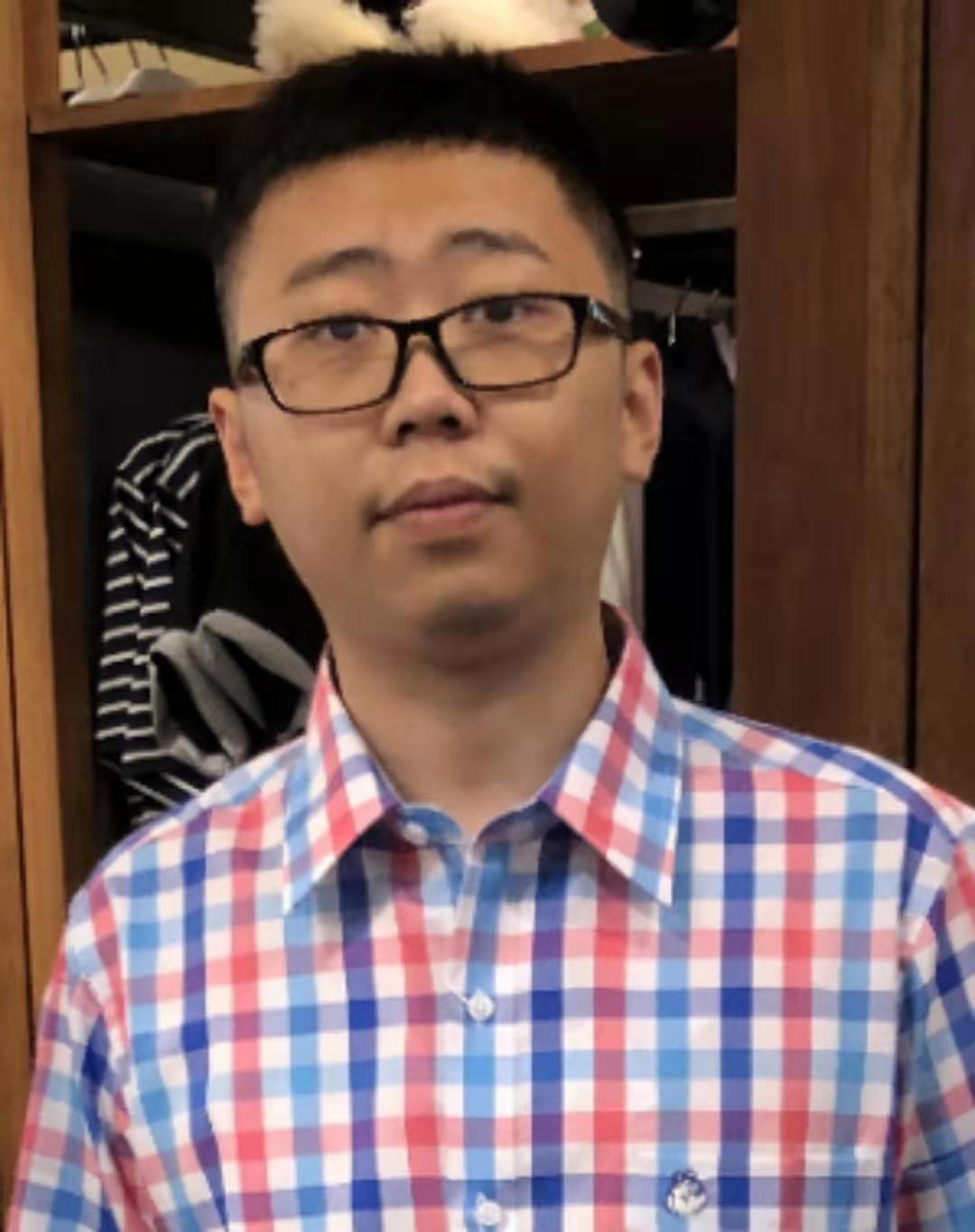}}]
{Rui Su} received the B.S. degree in school of Electronic and Information Engineering from South China University of Technology in 2013, and the MPhil. degree in school of Information Technology and Electrical Engineering from the University of Queensland in 2016. He is currently working toward the PhD. degree in the school of Electrical and Information Engineering, the University of Sydney. His current research interests include action detection and its applications in computer vision.
\end{IEEEbiography}

\begin{IEEEbiography}
[{\includegraphics[width=1in,height=1.25in,clip,keepaspectratio]{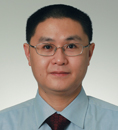}}]
{Dong Xu} received the BE and PhD degrees from University of Science and Technology of China, in 2001 and 2005, respectively. While pursuing the PhD degree, he was an intern with Microsoft Research Asia, Beijing, China, and a research assistant with the Chinese University of Hong Kong, Shatin, Hong Kong, for more than two years. He was a post-doctoral research scientist with Columbia University, New York, NY, for one year. He worked as a faculty member with Nanyang Technological University, Singapore. Currently, he is a professor and chair in Computer Engineering with the School of Electrical and Information Engineering, the University of Sydney, Australia. His current research interests include computer vision, statistical learning, and multimedia content analysis. He was the co-author of a paper that won the Best Student Paper award in the IEEE Conference
on Computer Vision and Pattern Recognition (CVPR) in 2010, and a paper that won the Prize Paper award in IEEE Transactions on Multimedia (T-MM) in 2014. He is a fellow of the IEEE.
\end{IEEEbiography}

\begin{IEEEbiography}
[{\includegraphics[width=1in,height=1.25in,clip,keepaspectratio]{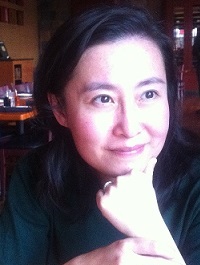}}]
{Luping Zhou} s now a Senior Lecturer in School of Electrical and Information Engineering at the University of Sydney, Australia. She received her PhD from Australian National University, and she was a recipient of Australian Research Council DECRA award (Discovery Early Career Researcher Award) in 2015. Her research interests include machine learning, computer vision, and medical image analysis. She is a senior member of the IEEE.
\end{IEEEbiography}

% \vspace{-140mm}
\begin{IEEEbiography}
[{\includegraphics[width=1in,height=1.25in,clip,keepaspectratio]{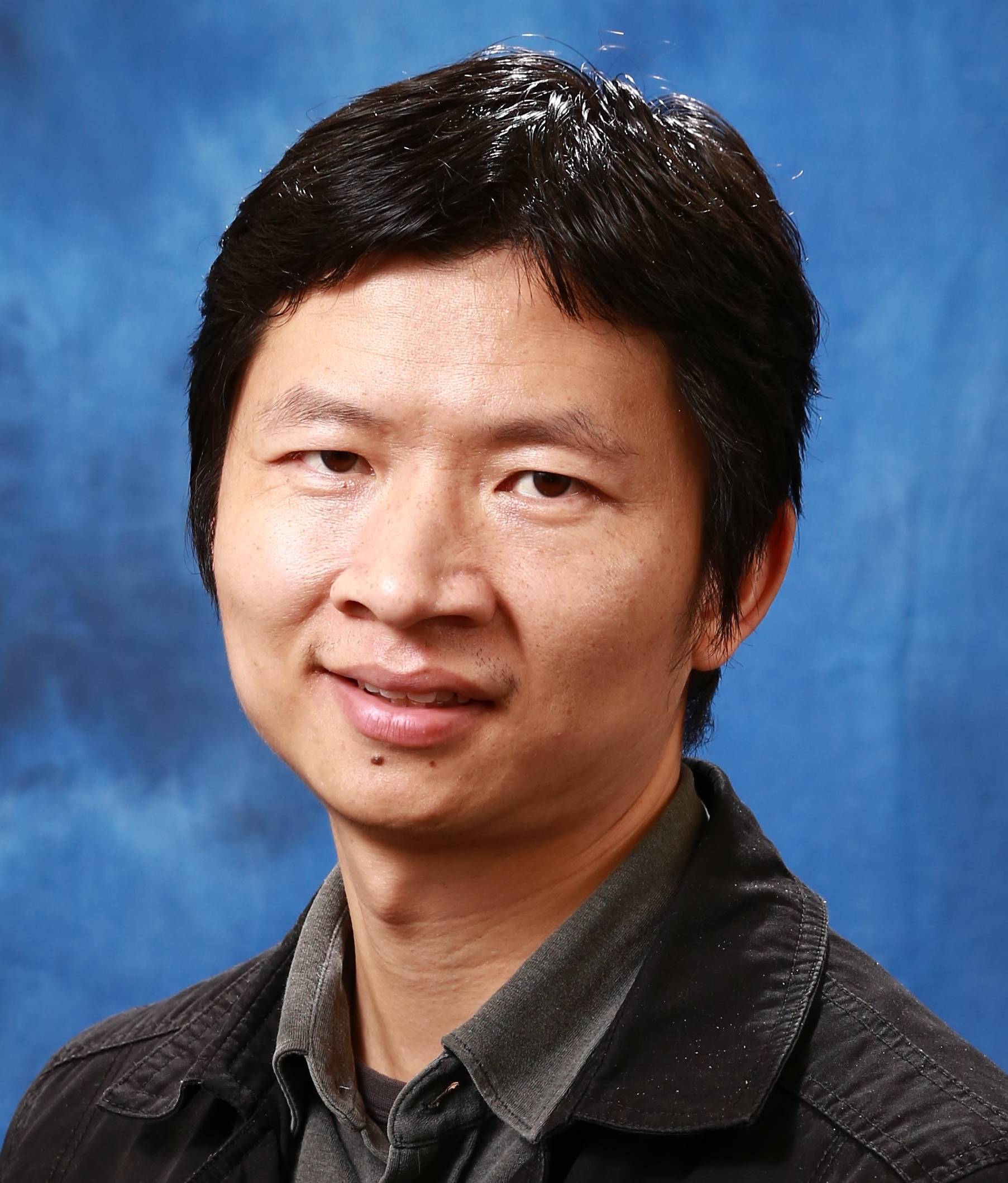}}]
{Wanli Ouyang} received the PhD degree in the Department of Electronic Engineering, Chinese University of Hong Kong. Since 2017, he is a senior lecturer with the University of Sydney. His research interests include image processing, computer vision, and pattern recognition. He is a senior member of the IEEE.
\end{IEEEbiography}


%\newpage



%\vfill


\end{document}